\documentclass[conference]{IEEEtran}

\usepackage{cite}
\usepackage{amsmath,amssymb,amsfonts}
\usepackage{algorithmic}
\usepackage{algorithm}
\usepackage{graphicx}
\usepackage{textcomp}
\usepackage{xcolor}
\usepackage{booktabs}
\usepackage{multirow}
\newcommand{\eg}{\emph{e.g.}}

\newcommand{\etal}{\emph{et al.}}

\usepackage[capitalize]{cleveref}
\def\BibTeX{{\rm B\kern-.05em{\sc i\kern-.025em b}\kern-.08em
    T\kern-.1667em\lower.7ex\hbox{E}\kern-.125emX}}
\begin{document}

\title{Model Inversion Attack Against Deep Hashing}

\author{
Dongdong Zhao\textsuperscript{1}, 
Qiben Xu\textsuperscript{1}, 
Ranxin Fang\textsuperscript{1}, 
Baogang Song\textsuperscript{1*}\\[0.5ex]
\textsuperscript{1}Wuhan University of Technology, China \\[0.5ex]
zdd@whut.edu.cn, xqb@whut.edu.cn, star\_fang0916@whut.edu.cn, 297710@whut.edu.cn\\[0.5ex]
}

\maketitle

\begingroup
\renewcommand{\thefootnote}{*}
\footnotetext{Corresponding author}
\endgroup


\begin{abstract}
Deep hashing improves retrieval efficiency through compact binary codes, yet it introduces severe and often overlooked privacy risks. The ability to reconstruct original training data from hash codes could lead to serious threats such as biometric forgery and privacy breaches. However, model inversion attacks specifically targeting deep hashing models remain unexplored, leaving their security implications unexamined. This research gap stems from the inaccessibility of genuine training hash codes and the highly discrete Hamming space, which prevents existing methods from adapting to deep hashing. To address these challenges, we propose DHMI, the first diffusion-based model inversion framework designed for deep hashing. DHMI first clusters an auxiliary dataset to derive semantic hash centers as surrogate anchors. It then introduces a surrogate-guided denoising optimization method that leverages a novel attack metric (fusing classification consistency and hash proximity) to dynamically select candidate samples. A cluster of surrogate models guides the refinement of these candidates, ensuring the generation of high-fidelity and semantically consistent images. Experiments on multiple datasets demonstrate that DHMI successfully reconstructs high-resolution, high-quality images even under the most challenging black-box setting, where no training hash codes are available. Our method outperforms the existing state-of-the-art model inversion attacks in black-box scenarios, confirming both its practical efficacy and the critical privacy risks inherent in deep hashing systems.
\end{abstract}


\section{Introduction}
\label{sec:intro}

With the massive growth of data, deep hashing \cite{Xia2014_2} has become a cornerstone for large-scale image retrieval. These methods map high-dimensional data into compact binary codes, enabling efficient storage and computation while preserving semantic similarity. Their deployment spans from efficiency-critical applications like image retrieval \cite{Liu2016_3, Chen2025_4} and real-time cross-camera tracking \cite{Zhu2017_5, Chen2017_6} to privacy-sensitive domains such as medical imaging \cite{Chen2024_7} and biometric recognition (\eg, face recognition systems \cite{Ghasemi2024_8}) \cite{Borra2024_9}. However, this efficient “high-dimensional $\to$ binary" compression introduces severe but often-overlooked privacy leakage risks as a trade-off for its performance gains.

Current research focuses on attacks that manipulate model outputs, such as backdoor implantation~\cite{backdoor_deephash} or hash collision generation~\cite{gen_collision_attack_deephash}. This focus has overshadowed the covert threat of Model Inversion Attacks (MI)~\cite{Fredrikson2014_10}, which aim to reconstruct either specific training instances or representative samples that capture the underlying data distribution from model outputs. The consequences of a successful MI on deep hashing could be severe. Reconstructed representative facial images may enable identity forgery~\cite{Kahla2022_32}, while medical model inversion can reveal specific genetic patterns of individuals~\cite{Subbanna2021_13}. These cases demonstrate that the exposure of hash codes poses a security risk by potentially revealing the underlying training data distribution.

\begin{figure}
  \centering
  \includegraphics[width=1\linewidth]{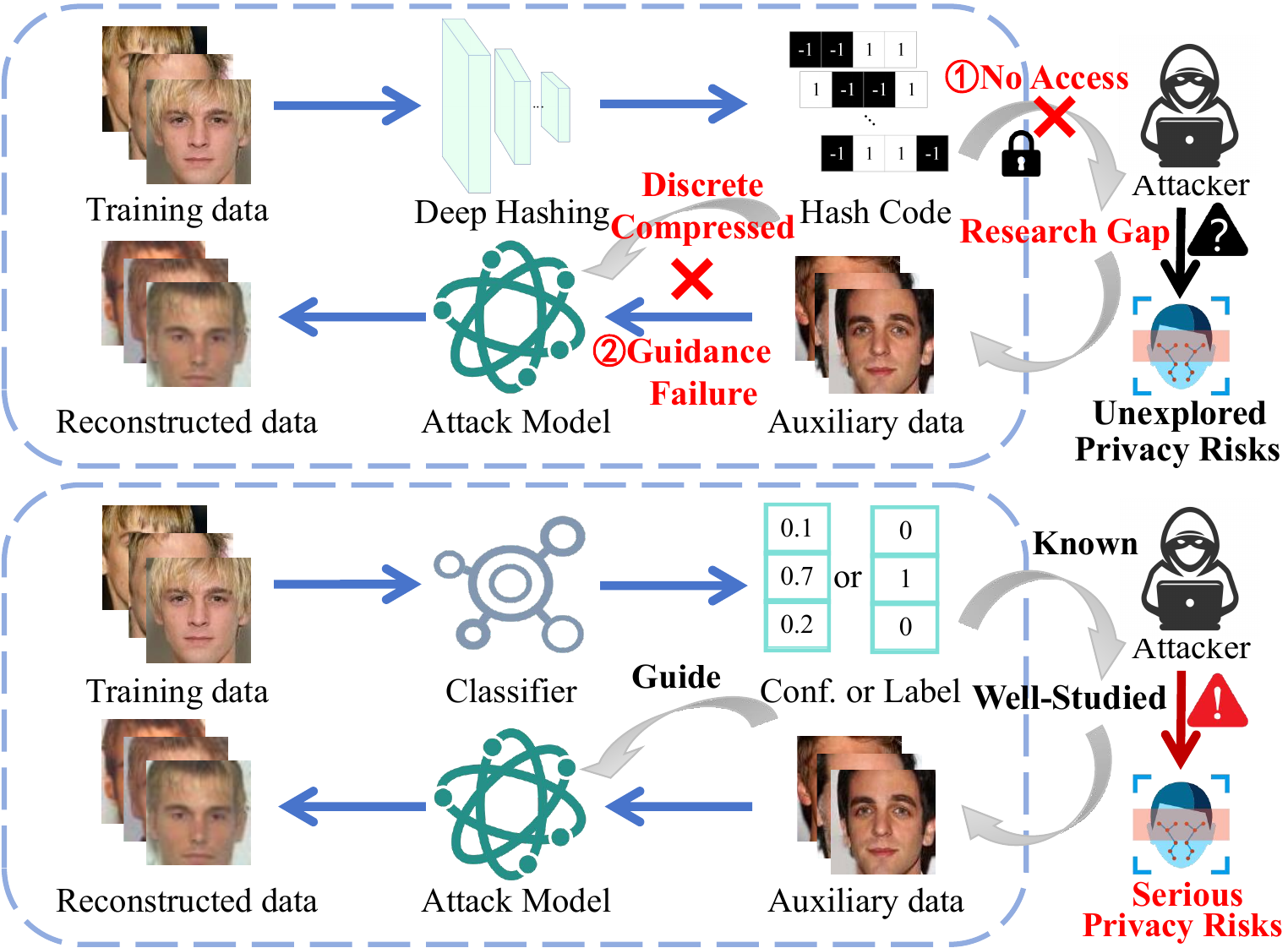}
  \caption{Research Motivation}
   \label{fig:research_motivation}
\end{figure}

However, existing model inversion attacks rely on a critical assumption: attackers have access to the target model's internal information, such as gradients and loss values, or model outputs on training data like confidence vectors and explicit semantic labels. In practice, deep hashing systems operate under a strict black-box setting where none of these prerequisites are available. As illustrated in~\cref{fig:research_motivation}, an attacker cannot access the original training hash codes. More fundamentally, even if the hash codes were obtainable, their discrete and highly compressed representation lacks the explicit semantic structure that guides existing inversion methods. This fundamental mismatch not only prevents the direct application of established attacks but also creates a significant and unexplored research gap in the security of deep hashing models.

To address the dual challenges of absent hash code supervision and the discrete, compressed Hamming space, we propose \textbf{DHMI}. It is the first diffusion-based model inversion framework specifically designed for deep hashing, and it operates without requiring any training hash codes. In general, the contributions of this study can be summarized as follows:

\begin{itemize}
\item We introduce DHMI, an effective inversion framework that, for the first time, addresses this task under a strict black-box setting with no training hash code access.

\item We propose a method to infer hash centers as semantic anchors from an auxiliary dataset to compensate for the lack of training hash codes.

\item We design a novel diffusion-based optimization strategy that ensures high-fidelity image generation while maintaining semantic consistency with the hash centers.

\item We demonstrate the superior performance of DHMI through extensive experiments on multiple datasets, where it significantly outperforms the existing state-of-the-art model inversion attacks.
\end{itemize}


\section{Related Works}
\label{sec:related_works}
\subsection{Model Inversion Attack}
\label{subsec:mi}

Model Inversion Attacks (MI) aim to extract characteristic information about the training data distribution, including reconstructing specific instances or representative features, from model outputs. Pioneered by Fredrikson \etal~\cite{Fredrikson2014_10, Fredrikson2015_11}, most subsequent deep learning-based MI methods target models with continuous, high-dimensional outputs (\eg, confidence scores). This creates a fundamental incompatibility with deep hashing, which produces discrete, compressed binary codes residing in a non-differentiable Hamming space that blocks gradient-based guidance and lacks explicit semantic distribution.

In white-box settings, attackers leverage full model access, typically relying on backpropagation through continuous output layers. Zhang \etal~\cite{Zhang2020_14} introduced Generative Model Inversion (GMI) using GANs, followed by variational frameworks (VMI~\cite{Wang2021_25}) and knowledge distillation techniques (KED-MI~\cite{Chen2021_26}) to enhance diversity and fidelity. Yuan \etal~\cite{Yuan2023_27} advanced this line with pseudo-label-guided conditional GANs. However, their core mechanism of gradient backpropagation is rendered ineffective by the non-differentiable nature of hash functions.

In black-box settings, attackers access only the model's final output. Methodologies have evolved from training inverse networks with auxiliary data~\cite{Yang2019_28} to optimization-based searches for optimal inputs \cite{Yoshimura2021_29, dionysiou2023_30}. Recent work DBB-MI~\cite{Bao2025_31} employs multi-agent reinforcement learning for distributional estimation. However, most of these methods require the specific outputs of the training samples to reconstruct them. Even DBB-MI~\cite{Bao2025_31}, which does not require training data information, still needs label information in addition to the confidence vector for guidance.

Under the most challenging label-only setting, attacks like BERPMI~\cite{Kahla2022_32} and LOKT~\cite{Nguyen2023_33} have improved reconstruction quality. Notably, \cite{Liu2024_15} leverages a conditional diffusion model~\cite{CDM} (CDM), using the explicit class label as a strong semantic guide. This highlights the fundamental obstacle for hashing-based MI: a hash code is not an interpretable label, creating a semantic gap that precludes the direct application of these techniques.


\subsection{Deep Hashing}
\label{subsec:deep_hash}
Deep hashing employs deep neural networks to map data into compact binary hash codes. A key characteristic across training paradigms is the implicit or explicit clustering of codes around specific centers in the Hamming space. While this clustering enhances retrieval efficiency, it also unveils a critical attack surface for privacy leakage, as these centers can be exploited to guide model inversion.


\subsubsection{Deep Supervised Hashing}
Supervised methods leverage semantic labels to learn discriminative hash functions, promoting clustering around semantic centers. Pairwise~\cite{Xia2014_2, Li2017_16} and ranking-based~\cite{Wang2017_17} methods preserve similarity structures, indirectly inducing clustering. Pointwise methods make this tendency explicit: some early works \cite{Jain2017_18} used classification layers to constrain code distribution. Furthermore, recent methods like CSQ~\cite{Yuan2020_19} and min-margin centers~\cite{Wang2023_20} directly optimize fixed, evenly-distributed hash centers to maximize separation. This paradigm of center-learning is also central to quantization-based methods, such as in~\cite{Klein2019_21}, which employs a codebook to represent the centers. Thus, cluster formation around semantic centers is a cornerstone of supervised deep hashing.


\subsubsection{Deep Unsupervised Hashing}
Without labels, unsupervised methods achieve similar goals by mining intrinsic data structures, forming implicit “cluster centers". Similarity reconstruction-based methods~\cite{Yang2018_22} preserve pairwise relationships, which promotes the formation of code clusters. Pseudo-label-based methods~\cite{Hu2017_23} introduce supervision into unsupervised hashing by utilizing cluster assignments as pseudo-labels to guide the learning of discriminative hash codes. Self-supervised methods like HashGAN~\cite{Dizaji2018_24} use adversarial training to learn cluster-preserving binary codes without supervision.


In summary, while model inversion is mature for continuous-output models, its strategies are mismatched for deep hashing due to the non-differentiable Hamming space and the loss of explicit semantics. This work bridges the gap by exploiting a universal phenomenon in deep hashing—the inherent clustering of codes—to pioneer practical model inversion without requiring original training hash codes.

\section{Threat Model}
\label{sec:threat_model}
This work investigates a realistic model inversion threat against deep hashing systems, focusing on a challenging black-box scenario. In this scenario, the attacker can only obtain binary hash codes via querying, with no access to the original training hash codes.


\begin{figure*}
  \centering
  \includegraphics[width=0.9\linewidth]{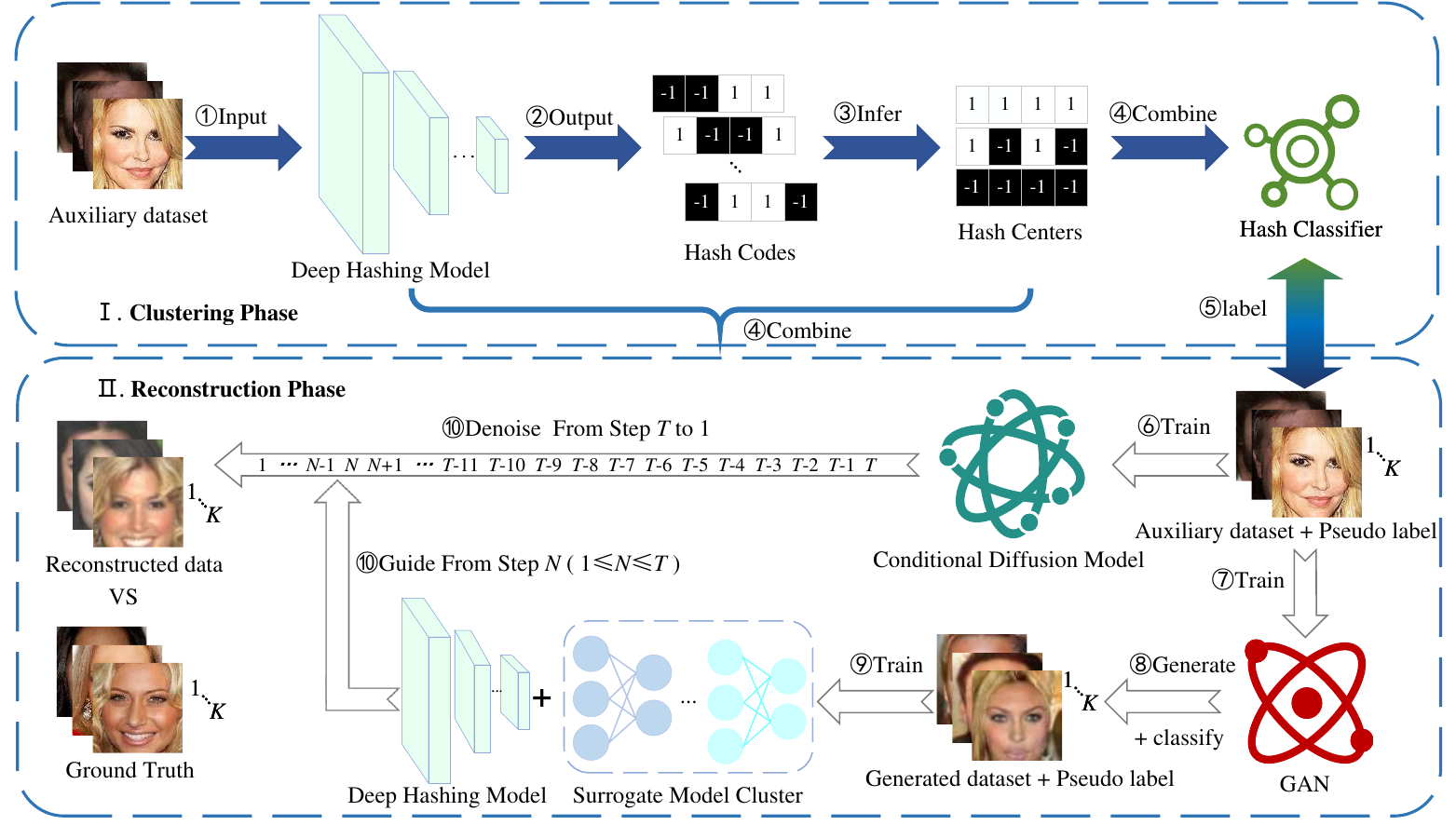}
  \caption{An Illustration of the Proposed Framework}
   \label{fig:framework}
\end{figure*}

\subsection{Attack Goals and Capabilities}
\label{subsec:attack_goals}

The attacker aims to reconstruct high-fidelity, semantically meaningful images from the private training set. The attacker's capabilities are confined to a query interface of the victim model, which provides only binary hash outputs for a given input. This severe information limitation makes inverting deep hashing models particularly challenging.


\subsection{Attacker Knowledge}
\label{subsec:attacker_knowledge}

This study adopts a set of practical and realistic assumptions regarding the attacker's prior knowledge:

\textbf{Task Knowledge:} The attacker knows that $F_H$ is a deep hashing model, which maps images to binary codes. These assumptions are reasonable given the public nature of the model’s task.

\textbf{Data Knowledge:} The attacker has access to a public auxiliary dataset $\mathcal{D}_{aux}$ that shares the same data distribution as $F_H$'s training data $\mathcal{D}_{priv}$, but has non-overlapping classes. For instance, $\mathcal{D}_{priv}$ could be a private celebrity dataset, while $\mathcal{D}_{aux}$ comprises web-crawled face images.
    
\textbf{Model Knowledge:} The attacker operates under a black-box setting, without any knowledge of the model's architecture or parameters, and can only interact with it solely by querying a provided API.


\subsection{Problem Formulation}
\label{subsec:problem_formulation}
The attacker interacts with the target deep hashing model $F_H: \mathcal{X} \rightarrow \{-1, +1\}^l$ by providing input images $x$ and observing the resulting binary hash codes $h$. Formally, this interaction is defined by the constraint $F_H(x) = h$, where $x \in \mathcal{X}$ is a query image and $h \in \{-1, +1\}^l$ is its corresponding $l$-bit binary hash code.

Given the threat model above, the model inversion attack against $F_H$ aims to reconstruct high-fidelity, semantically meaningful images from $\mathcal{D}_{priv}$. To achieve this goal, the attacker needs to solve two fundamental problems:
The first problem is to infer a set of semantic hash centers ${C} = \{h_{1}, \dots, h_{K}\}$ for the $K$ classes in $\mathcal{D}_{priv}$. It can be formulated as finding the optimal center $h_{i}^*$ for each cluster $\mathcal{Q}_{i}$ around the genuine distribution centers by solving:
\begin{equation}
h_i^* = \arg\min_{h \in \{-1, +1\}^l} \sum_{x \in \mathcal{Q}_i} \mathcal{D}_H\left(F_H(x), h\right),\forall i \in \{1, \dots, K\}
\label{eq:single_center_optimization}
\end{equation}
where $\mathcal{D}_H(\cdot, \cdot)$ is the Hamming distance metric.

The second problem is, for each inferred hash center $h_i \in {C}, \forall i \in \{1, ..., K\}$, to find the optimal reconstructed image $\tilde{x}_i^*$ that minimizes the Hamming distance between the model's output and the target center:
\begin{equation}
\label{eq:problem_formulation}
\tilde{x}_{i}^* = \underset{\tilde{x}\in \mathcal{X}}{\arg\min} \, \mathcal{D}_H\left( F_H({\tilde{x}}), \, h_{i} \right), \forall i \in \{1, ..., K\}
\end{equation}

These core problems are challenged by three key constraints derived from the threat model: (1) the \textbf{unavailability} of the true training hash codes and their genuine distribution centers; (2) the \textbf{non-differentiability} of $F_H$ and the Hamming space, which prevents gradient-based optimization;  and (3) the need to search within a \textbf{semantically meaningful image manifold} to produce realistic outputs, as direct pixel-space optimization is ill-posed~\cite{Nguyen2023_33}.


\section{DHMI}
\label{sec:methodology}

We propose the Deep Hashing Model Inversion (DHMI) framework, a methodology designed to overcome the three fundamental barriers in deep hashing inversion: (1) the unavailability of training hash codes, compensated by \textbf{inferring semantic centers} from an auxiliary dataset; (2) the non-differentiable Hamming space, navigated via \textbf{a hybrid guidance strategy} that couples surrogate model gradients with direct target model queries; and (3) the need for high-fidelity image generation, achieved by leveraging \textbf{a conditional diffusion model} trained on the pseudo-labeled data.


\subsection{Overview of the Proposed Framework}
\label{subsec:overview}
As illustrated in~\cref{fig:framework}, the DHMI framework operates in two distinct phases:

\textbf{Phase 1: Clustering Phase.} This phase aims to infer the semantic structure of the target model $F_H$. We first \textbf{input} the auxiliary dataset $\mathcal{D}_{aux}$ into $F_H$ to obtain the \textbf{output} hash code matrix $H \in \{-1, +1\}^{n \times l}$ (where $n$ denotes the number of samples in $\mathcal{D}_{aux}$). We then \textbf{infer} the underlying semantic hash centers ${C}$ from $H$ using our proposed estimation method (\cref{subsec:Clustering}). Next, we \textbf{combine} the centers ${C}$ with $F_H$ to construct a hash classifier. Finally, this classifier is used to \textbf{label} $\mathcal{D}_{aux}$ with pseudo-labels, which guide the subsequent reconstruction phase.

\textbf{Phase 2: Reconstruction Phase.} This phase focuses on generating high-fidelity images semantically aligned with the hash centers ${C}$. We first \textbf{train} a conditional diffusion model on the pseudo-labeled $\mathcal{D}_{aux}$. To efficiently approximate the underlying distribution of the private training data as closely as possible, we propose a hybrid strategy which incorporates a prior technique~\cite{Nguyen2023_33}: we \textbf{train} a GAN to generate a large-scale, class-balanced dataset, \textbf{classify} it with our hash classifier, and \textbf{train} a surrogate model cluster on this generated dataset. The core of our method is \textbf{the guided denoising process} (\cref{subsec:Denoising})  from step $T$ to $1$. To safeguard the critical early stages of generation, external guidance is activated only from step $N (1 \leq N \leq T)$ onward. Throughout the subsequent denoising steps, we employ gradients from the surrogate model cluster alongside constraints from $F_H$ to iteratively optimize the reconstructed images, directing them towards ${C}$. This process ultimately yields high-fidelity images that exhibit both precise hash-code alignment and strong semantic consistency.

In summary, DHMI constitutes a systematic framework that addresses the core inversion challenges. Its design sequentially resolves the problem by first inferring the missing semantic structure and then leveraging it to steer the sampling process of a conditional diffusion model, thereby generating high-fidelity, semantically faithful reconstructions under strict black-box constraints.

\begin{algorithm}[htbp]
\caption{Slice-Fused Hash Center Estimation}
\label{alg:clustering}
\textbf{Input:} Hash code matrix $H \in \{-1, +1\}^{n \times l}$ of auxiliary dataset; number of hash centers $K$; base slice size $s_{base}$; neighborhood radius ratio $r$; slice overlap ratio $o$ \\
\textbf{Output:} Hash center matrix $C \in \{-1, +1\}^{K \times l}$\\
1: $C \leftarrow \text{K-means}(H, K)$ \\
2: $slices_{idx} \leftarrow \text{EmptyList}()$ \\
3: $step \leftarrow \max(1, \lfloor s_{base} \times (1 - o) \rfloor)$ \\
4: \textbf{for} $start = 0$ \textbf{to} $l-1$ \textbf{by} $step$ \textbf{do} \\
5: \hspace{0.5cm} $end \leftarrow \min(start + s_{base}, l)$ \\
6: \hspace{0.5cm} $idx \leftarrow [start, start+1, \dots, end-1]$ \\
7: \hspace{0.5cm} $slices_{idx} \leftarrow idx \cup slices_{idx}$\\
8: \textbf{end for} \\
9: \textbf{for} $i = 0$ \textbf{to} $K-1$ \textbf{do} \\
10: \hspace{0.35cm} $D \leftarrow \text{HammingDistance}(H, C[i])$ \\
11: \hspace{0.35cm} $radius \leftarrow \max(1, \lfloor l \cdot r \rfloor)$\\
12: \hspace{0.35cm} $neighbors \leftarrow \{ H_j \mid D_j \leq radius, j = 1,...,n\}$\\
13: \hspace{0.35cm} \textbf{for each} $idx$ \textbf{in} $slices_{idx}$ \textbf{do} \\
14: \hspace{0.85cm} $slices \leftarrow neighbors[idx]$ \\
15: \hspace{0.85cm} $slice\_count \leftarrow \text{EmptyDictionary}()$ \\
16: \hspace{0.85cm} \textbf{for each} $slice$ \textbf{in} $slices$ \textbf{do} \\
17: \hspace{1.35cm} $slice\_count[slice] \leftarrow slice\_count[slice] + 1$ \\
18: \hspace{0.85cm} \textbf{end for} \\
19: \hspace{0.85cm} $C[i][idx] \leftarrow \arg\max\limits_{slice} slice\_count[slice]$\\
20: \hspace{0.35cm} \textbf{end for} \\
21: \textbf{end for} \\
22: \textbf{return} $C$

\end{algorithm}

\subsection{Slice-Fused Hash Center Estimation}
\label{subsec:Clustering}
To infer the semantic structure of $F_H$, we propose a slice-fused method comprising two key steps. First, we apply K-means clustering~\cite{kmeans} to the hash codes of $\mathcal{D}_{aux}$ to obtain initial cluster centers. Second, since these initial centers may not adequately capture the true semantics of the private distribution, we refine them through local statistical aggregation. Specifically, for each initial center, we define its neighborhood as an approximation to the genuine cluster $\mathcal{Q}_i$ and update the center by selecting the most frequent bit patterns. This approach effectively transforms the optimization objective from direct Hamming distance minimization to an indirect frequency-based estimation, thereby better approximating the target distribution centers defined in ~\cref{eq:single_center_optimization}.

The complete workflow is detailed in~\cref{alg:clustering}. It first applies K-means clustering to the hash matrix $H$, obtaining initial cluster centers $C$ (Step 1). It then generates a sequence of overlapping slice indices (Steps 2-8) via a sliding window of size $s_{base}$ with overlap ratio $o$. This overlap design enhances robustness by ensuring bit positions are covered by multiple slices.

The core refinement iterates through each initial center (Step 9). For center $C[i]$, its neighbors in $H$ are defined as the set of points within a Hamming radius of $l \times r$ (Steps 10-12). For each slice index $idx$ in $slices_{idx}$ (Steps 13-20), it extracts neighbor bit patterns, computes their frequency distribution, and selects the most prevalent pattern to update $C[i]$ at position $idx$.

When multiple slices cover the same bit, subsequent assignments overwrite previous ones, preserving the latest local information. The final output is the refined discrete hash center matrix $C$, serving as high-quality semantic anchors for inversion attacks.
 

\subsection{Surrogate-Driven Image Denoising}
\label{subsec:Denoising}
To overcome the Hamming space's non-differentiability and the inefficiency of direct pixel-space optimization, we propose a novel diffusion-based fusion guidance strategy that integrates conditional generation with surrogate-model guidance and target model constraints.

Our approach builds upon the conditional diffusion model (CDM) framework, which consists of two complementary processes. The forward process progressively corrupts a clean image $x$ by adding Gaussian noise over $T$ steps, transforming it into pure noise $x_T \sim \mathcal{N}(\mathbf{0}, \mathbf{I})$. Conversely, the reverse process reconstructs a high-quality image $x_0$ by iteratively denoising from the Gaussian noise $x_T$ through $T$ sampling steps. The core training objective of the CDM is to learn a noise prediction model that approximates the noise added during the forward process, thereby enabling effective reverse denoising.

Following the CDM framework established in~\cite{Liu2024_15}, we employ a classifier-free guidance mechanism. The conditioned noise prediction is formulated as:
\begin{equation}
\tilde{\epsilon}_\theta(x_t, t, y) = (1 + \omega) \epsilon_\theta(x_t, t, y) - \omega \epsilon_\theta(x_t, t, \emptyset)
\label{eq:predict_noise}
\end{equation}
where $\epsilon_\theta$ represents the CDM, $y$ represents the pseudo-label assigned by the hash classifier, which is stochastically replaced with a null label $\emptyset$ during training with probability $p$ to enable unconditional guidance, $t$ represents the current denoising step, and $\omega$ controls the guidance strength.

The reverse denoising step is then computed as:
\begin{equation}
x_{t-1} = \frac{1}{\sqrt{\alpha_t}} \left( x_t - \frac{\beta_t}{\sqrt{1 - \bar{\alpha}_t}} \tilde\epsilon_\theta(x_t, t, y) \right) + \sqrt{{\beta}_t} \mathbf{z}
\label{eq:denoise}
\end{equation}
where $\bar{\alpha}_t = \prod_{i=1}^t \alpha_i$, $\alpha_t = 1 - \beta_t$, $\beta_t$ is the pre-defined noise variance at step $t$, and $\mathbf{z} \sim \mathcal{N}(\mathbf{0}, \mathbf{I})$.

We thus solve the pixel-space optimization problem by reframing it within the CDM's latent space, guided by its learned generative knowledge. For each target hash center $h_{i} \in {C}, \forall i \in \{1, ..., K\}$, in order to reveal as much of its underlying training data distribution as possible, we initiate the reconstruction process by sampling $n$ initial noise vectors $\{x_T^{(1)}, \ldots, x_T^{(n)}\}$ from $\mathcal{N}(0, \mathbf{I})$ and denoising them to obtain candidate images $\{x_0^{(1)}, \ldots, x_0^{(n)}\}$ (Steps 1-11). During denoising, we preserve the intermediate latent $x_{N-1}$ at step $N$ (Step 8) as the starting point for refinement.

To select candidates that are both fusing classification consistency and hash proximity, we design an attack adaptation metric $S_{attack}$ that combines the stability of the hash classifier's predictions under data augmentations with the proximity of the hash code to the target center $h_i$:
\begin{equation}
S_{attack}(x_0) = \frac{w_{{base}} \cdot d_{i0} \cdot M \cdot \delta(\hat{y_0}, y) + \sum_{t=1}^{M} d_{it} \cdot \delta(\hat{y_t}, y)}{w_{{base}} \cdot d_{i0} \cdot M \cdot \delta(\hat{y_0}, y) + M}
\label{eq:robustness_score}
\end{equation}
Here, $\delta$ is an indicator function that returns 1 when the predicted label $\hat{y}_t$ after the $t$-th transformation equals the target label $y$ and 0 otherwise. The target label $y$ corresponds to the pseudo-label assigned to the target hash center $h_i$ during the clustering phase; $w_{{base}}$ is the initial prediction weight; $M$ is the number of data augmentations; and $d_{ij}$ is the Hamming distance-based weight between the image $x_j$'s hash code and the target hash center $h_i$, can be calculated as:
\begin{equation}
d_{ij} = \exp\left(-\frac{dis_{ij}}{l} \cdot w_{{hamming}}\right)
\label{eq:hamming_weight}
\end{equation}

\begin{algorithm}[htbp]
\caption{Surrogate-Guided Diffusion Inversion}
\label{alg:denoise}
\textbf{Input:} Conditional diffusion model $\epsilon_{\theta}$; target label $y$; guidance weight $\omega$; total noise steps $T$; Optimize times $iter$; learning rate $lr$; Intermediate denoising steps $N$\\
\textbf{Output:} Optimized images set $\mathcal{\tilde{X}}:\{\tilde{x}^{(1)}, \ldots, \tilde{x}^{(k)}\}$ \\
1: Generate $n$ noise images: $\{{x}_T^{(1)}, \ldots, {x}_T^{(n)}\} \sim \mathcal{N}(0, \mathbf{I})$ \\
2: $\mathcal{X}_{N-1} \leftarrow \emptyset$, $\mathcal{X}_0 \leftarrow \emptyset$, $\mathcal{\tilde{X}} \leftarrow \emptyset$\\
3: \textbf{for} $i = 1$ \textbf{to} $n$ \textbf{do} \\
4: \hspace{0.5cm} \textbf{for} $t = T$ \textbf{down to} $1$ \textbf{do} \\
5: \hspace{1.0cm} calculate $\tilde{\epsilon}_\theta({x_t}^{(i)}, t, y)$ according to \ref{eq:predict_noise} \\
6: \hspace{1.0cm} $\mathbf{z} \sim \mathcal{N}(\mathbf{0}, \mathbf{I})$ \textbf{if} $t > 1$, \textbf{else} $\mathbf{z} = 0$ \\
7: \hspace{1.0cm} calculate $x_{t-1}$ according to \ref{eq:denoise}\\
8: \hspace{1.0cm} \textbf{if} $t = N$ \textbf{then} $\mathcal{X}_{N-1} \leftarrow x_{t-1} \cup \mathcal{X}_{N-1}$ \\
9: \hspace{0.50cm} \textbf{end for} \\
10: \hspace{0.35cm} $\mathcal{X}_0 \leftarrow x_0 \cup \mathcal{X}_0$ \\
11: \textbf{end for} \\
12: $\{{x}_{0}^{(1)}, \ldots, {x}_{0}^{(k)}\} \leftarrow \text{Top-k}(S_{attack}(\mathcal{X}_0))$ \\
13: \textbf{for} $i = 1$ \textbf{to} $k$ \textbf{do} \\
14: \hspace{0.35cm} $\tilde{x} \leftarrow {x}_{0}^{(i)}$ \\
15: \hspace{0.35cm} \textbf{for} $j = 1$ \textbf{to} $iter$ \textbf{do} \\
16: \hspace{0.85cm} ${x}_t \leftarrow$ Load saved state ${{x}_{N-1}}^{(i)}$ of $x_0^{(i)}$ \\
17: \hspace{0.85cm} \textbf{for} $t = N - 1$ \textbf{down to} $1$ \textbf{do} \\
18: \hspace{1.35cm} calculate $\tilde{\epsilon}_\theta({x_t}, t, y)$ according to \ref{eq:predict_noise} \\
19: \hspace{1.35cm} $\mathbf{z} \sim \mathcal{N}(\mathbf{0}, \mathbf{I})$ \textbf{if} $t > 1$, \textbf{else} $\mathbf{z} = 0$ \\
20: \hspace{1.35cm} calculate $x_{t-1}$ according to \ref{eq:denoise}\\
21: \hspace{1.35cm} add ${x}_{t-1}$ to Adam optimizer with $lr$\\
22: \hspace{1.35cm} calculate $\mathcal{L}_{sur}$ according to \ref{eq:surrogate_loss}\\
23: \hspace{1.35cm} optimize ${x}_{t-1}$ with Adam to minimize $\mathcal{L}_{sur}$ \\
24: \hspace{0.85cm} \textbf{end for} \\
25: \hspace{0.85cm} \textbf{if} $S_{attack}(x_0) > S_{attack}(\tilde{x})$ \textbf{then} $\tilde{x} \leftarrow {x}_{0}$ \\
26: \hspace{0.35cm} \textbf{end for} \\
27: \hspace{0.35cm} $\mathcal{\tilde{X}} \leftarrow \tilde{x} \cup \mathcal{\tilde{X}}$\\
28: \textbf{end for} \\
29: \textbf{return} $\mathcal{\tilde{X}}:\{\tilde{x}^{(1)}, \ldots, \tilde{x}^{(k)}\}$
\end{algorithm}

We thus reframe the problem (Eq.~\eqref{eq:problem_formulation}) for set-based distribution leakage and select the top-$k$ images using the $S_{attack}$ metric (Step 12):
\begin{equation}
{x}_{0}^{(j)*} = \arg\max_{{x}_0^{(i)} \in {x}_0^{(n)}} S_{attack}(x_0^{(i)}), \forall j \in \{1, ..., k\}
\label{eq:topk_selection}
\end{equation}

While the selected images are semantically consistent, they lack precise hash code alignment. We therefore refine them by optimizing their preserved latent states $x_{N-1}$ using a surrogate model cluster for $iter$ iterations (Steps 13-27), bypassing the need for diffusion model gradients.

During each iteration, we denoising from $x_{N-1}$ to $x_0$ (Steps 17-24), incorporating each $x_{t-1}$ into the Adam optimizer. The surrogate loss is computed as:
\begin{equation}
\mathcal{L}_{{sur}} = \frac{1}{m} \sum_{i=1}^{m} \mathcal{L}_{\mathrm{CE}}(S_i(x_{t-1}), y)
\label{eq:surrogate_loss}
\end{equation}
where $S_i(x_{t-1})$ represents the probability output of the $i$-th surrogate model for image $x_{t-1}$. We optimize $x_{t-1}$ to minimize $\mathcal{L}_{sur}$, steering generation toward target semantics.

After each refinement, we evaluate the candidate using $S_{attack}$. If the metric increases, we update the current best candidate; otherwise, we retain the previous version (Step 25). After $iter$ iterations, the top-$k$ candidates with the highest $S_{{attack}}$ are selected as the final reconstructed images for the hash center $h_{i}$. The complete optimization procedure is summarized in~\cref{alg:denoise}.


\section{Experiments}
\label{sec:experiments}

\subsection{Experimental Setup}
\label{subsec:experimental_setup}

\subsubsection{Datasets}
\label{subsubsec:datasets}
We evaluate our method on three face recognition datasets commonly used in model inversion attacks: CelebA~\cite{Celeba}, FaceScrub~\cite{FaceScrub}, and PubFig83~\cite{Pubfig83}. Following the established setting in~\cite{Nguyen2023_33}, we split each dataset into two subsets with non-overlapping identities: a private set \(\mathcal{D}_{priv}\) for training $F_H$, and a public auxiliary set \(\mathcal{D}_{aux}\) used by the attacker. To further examine the robustness of our approach under auxiliary data distribution shift, we incorporate the FFHQ~\cite{ffhq} dataset as an additional \(\mathcal{D}_{aux}\) source. All face images are centrally cropped and resized to \(64 \times 64\) pixels to minimize the influence of background context.
\subsubsection{Models}
\label{subsubsec:models}

Our experimental setup uses ResNet-50~\cite{resnet} backbone for the target deep hashing models, implemented following~\cite{Wang2023_20}. For the attack component, our DHMI framework utilizes the CDM from CDMMI~\cite{Liu2024_15} trained on $\mathcal{D}_{\text{aux}}$. The surrogate model cluster consists of three DenseNet~\cite{Densenet} architectures (DenseNet-121, DenseNet-161, and DenseNet-169), trained consistently with LOKT~\cite{Nguyen2023_33}. For evaluation, we assess attack accuracy using FaceNet~\cite{FaceNet} and VGG16~\cite{VGG16}, and measure retrieval performance using mean average precision (mAP) with a ResNet-101~\cite{resnet} backbone featuring different hash centers, trained according to~\cite{Wang2023_20}. All evaluation models are trained on the same data as the target models. The mAP results of the target model on the three datasets are presented in~\cref{tab:map_of_resnet50}.



\begin{table}[htbp]
  \centering
  \caption{Mean Average Precision on Target Model(MAP@all)}
  \label{tab:map_of_resnet50}
  \begin{tabular}{lcccc}
    \toprule {Dataset}& 16bits & 32bits & 64bits & 128bits \\
    \midrule
    CelebA    & 78.12\% & 82.46\% & 83.91\% & 87.35\% \\
    FaceScrub & 78.83\% & 82.88\% & 84.28\% & 86.99\% \\
    Pubfig83  & 80.80\% & 83.76\% & 86.22\% & 87.72\% \\
    \bottomrule
  \end{tabular}
\end{table}

\begin{table*}[htbp]
  \centering
  \small 
  \setlength{\tabcolsep}{3.5pt} 
  \caption{Mean Hamming Distance After Alignment with Ground Truth}
  \label{tab:distance_of_predict_centers}
  \begin{tabular}{cccccccccccccc}
    \toprule\multirow{2}{*}{Dataset} & \multicolumn{3}{c}{16bits} & \multicolumn{3}{c}{32bits} & \multicolumn{3}{c}{64bits} & \multicolumn{3}{c}{128bits}\\
    \cmidrule(lr){2-4} \cmidrule(lr){5-7} \cmidrule(lr){8-10} \cmidrule(lr){11-13}
    & Random & K-means & Ours & Random & K-means & Ours & Random & K-means & Ours & Random & K-means & Ours & \\
    \midrule
    CelebA    & 3.40 & 1.33 & \textbf{1.23} & 9.49 & 2.01 & \textbf{1.73} & 22.75 & 2.90 & \textbf{2.41} & 51.00 & 10.69 & \textbf{7.91} \\
    FaceScrub & 2.95 & 0.99 & \textbf{0.97} & 8.88 & 1.21 & \textbf{1.01} & 21.57 & 2.56 & \textbf{1.96} & 49.17 & 7.85 & \textbf{6.46} \\
    Pubfig83  & 3.88 & 0.58 & \textbf{0.54} & 10.14 & 1.44 & \textbf{1.26} & 23.68 & 3.42 & \textbf{2.98} & 52.74 & 7.86 & \textbf{7.00} \\
    FFHQ $\to$ Pubfig83  & 3.80 & 0.58 & \textbf{0.56} & 10.10 & 1.94 & \textbf{1.92} & 23.58 & 4.44 & \textbf{4.26} & 52.16 & 12.44 & \textbf{10.76} \\
    \bottomrule
  \end{tabular}
\end{table*}

\begin{table*}[htbp]
  \centering
  \small 
  \setlength{\tabcolsep}{2.5pt} 
  \caption{Comparison with Existing Methods (Slice-Fused, 64bits)}
  \label{tab:acc_of_predict_centers}
  \begin{tabular}{cccccccccccccc}
    \toprule\multirow{2}{*}{Method} & \multicolumn{3}{c}{CelebA} & \multicolumn{3}{c}{FaceScrub} & \multicolumn{3}{c}{Pubfig83} &
    \multicolumn{3}{c}{FFHQ $\to$ Pubfig83}\\
    \cmidrule(lr){2-4} \cmidrule(lr){5-7} \cmidrule(lr){8-10} \cmidrule(lr){11-13}
    & Top-1 $\uparrow$ & Top-5$\uparrow$ & KNN dist.$\downarrow$ & Top-1 $\uparrow$ & Top-5 $\uparrow$ & KNN dist.$\downarrow$ & Top-1 $\uparrow$ & Top-5 $\uparrow$ & KNN dist.$\downarrow$ & Top-1 $\uparrow$ & Top-5 $\uparrow$ & KNN dist.$\downarrow$\\
    \midrule
    BERPMI & 40.60\%& 62.00\% & 1313.38 & 48.00\% & 74.00\% & 2004.88 & 34.40\% & 58.00\% & 1098.52 & 20.00\% & 44.00\% & 1103.41\\
    LOKT & 70.00\% & 86.80\% & 1089.51 & \textbf{90.40}\% & \textbf{95.00\%} & \textbf{1594.24} & 68.00\% & 81.60\% & 830.62 & 33.60\% & 66.00\% & 1029.54\\
    CDMMI & 51.00\% & 81.00\% & 1198.12 & 60.80\% & 85.80\% & 1804.63 & 48.80\% & 74.40\% & 926.43 &42.80\% & 74.80\% &
    1013.92\\
    Ours & \textbf{79.60\%} & \textbf{88.40\%} & \textbf{1086.26} & \underline{88.80}\% & \textbf{95.00\%} & \underline{1653.46} & \textbf{73.60\%} & \textbf{83.60}\% & \textbf{817.82} & \textbf{56.00}\% & \textbf{81.20}\% & \textbf{973.05} \\
    \bottomrule
  \end{tabular}
\end{table*}

\subsubsection{Implementation Details}
\label{subsubsec:implementation_details}
To ensure fair comparison and generality, we determined a set of hyperparameters that are fixed across all datasets through experiments balancing attack accuracy and high fidelity. The specific configurations are as follows: the number of data augmentations \(M = 300\), the surrogate model cluster size \(m = 3\), the number of denoising steps \(N = 100\), the initial candidate pool size \(n = 40\), the number of selected candidates \(k = 5\), the base prediction weight \(w_{{base}} = 0.2\), the Hamming weight scaling factor \(w_{{hamming}} = 5\), the number of optimization iterations \(\mathit{iter} = 6\), and the Adam optimizer's learning rate \(\mathit{lr} = 0.0015\).

All experiments were conducted on an NVIDIA GeForce RTX 4090 GPU with 24 GB of memory. Using CelebA as the auxiliary dataset under our training setup, the attack model completed one round of training in approximately 4 minutes, generated 40 images per category in 5 minutes, and finished one optimization iteration over the final 100 denoising steps for 5 selected images in just 9 seconds.


\subsubsection{Attack Comparison and Evaluation}
\label{subsubsec:attack_eval}
With no existing methods for deep hashing inversion, we compare three label-only approaches that can use our estimated centers: BERPMI~\cite{Kahla2022_32}, LOKT~\cite{Nguyen2023_33}, and CDMMI~\cite{Liu2024_15}. All methods are evaluated identically, testing our centers against K-means and random baselines.

We employ three evaluation metrics: \textbf{Attack Accuracy (Attack Acc)} measures target identity recognition performance; \textbf{K-Nearest Neighbor Distance (KNN Dist)} computes the minimal feature distance to private training images; and \textbf{Mean Average Precision (mAP)} evaluates retrieval performance in hashing tasks.

\subsection{Comparative Experiments}
\label{subsec:comparative_xperiments}
\begin{figure}
  \centering
  \includegraphics[width=0.9\linewidth]{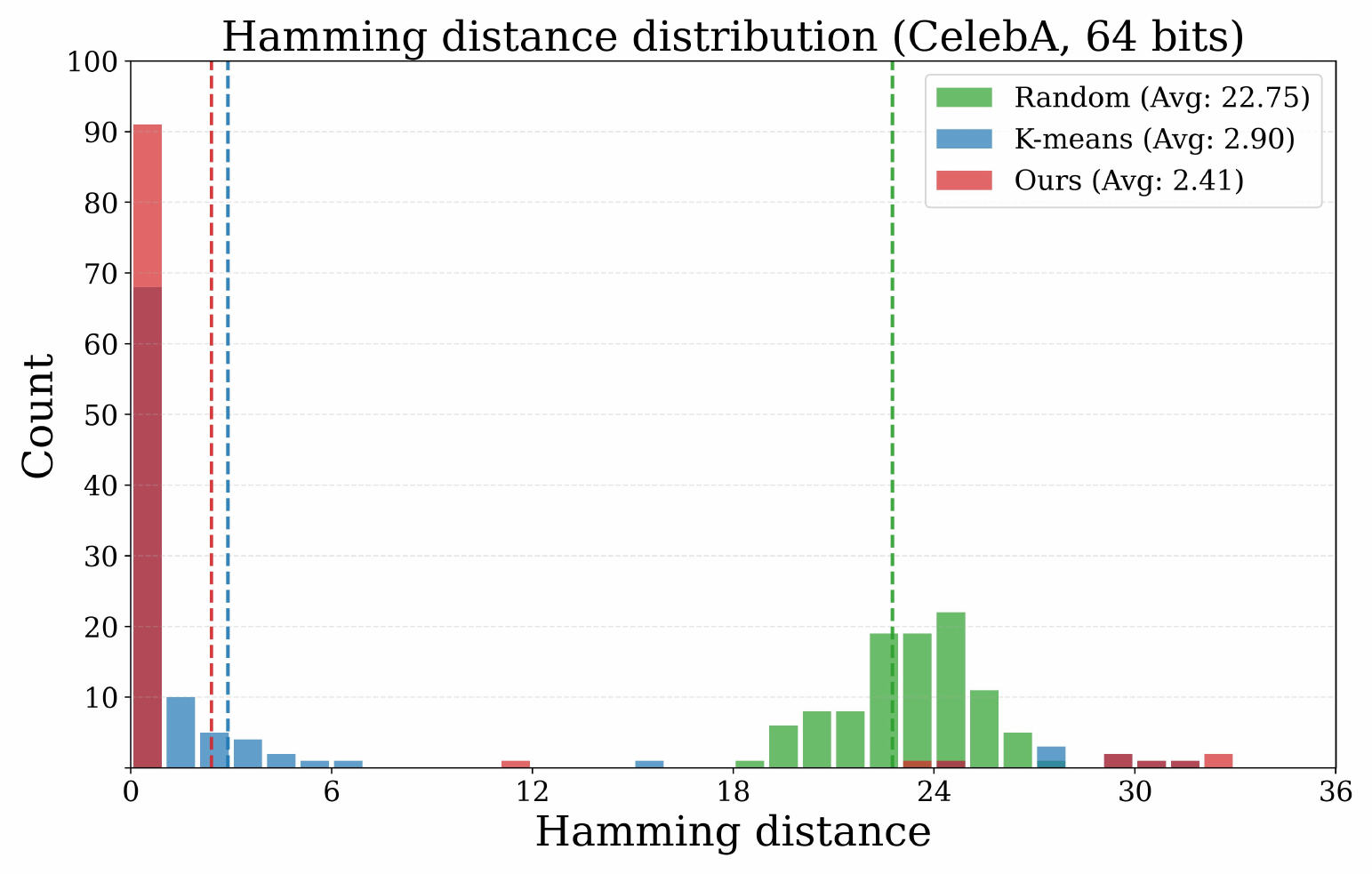}
  \caption{Distribution of Hamming Distances After Alignment}
   \label{fig:distribution}
\end{figure}
\subsubsection{Comparison of Hash Center Prediction Methods}
\label{subsubsec:compare_hash_center}
We evaluate our slice-fused hash center estimation against two baselines: random and K-means centers. For fair comparison, all methods use the Hungarian algorithm~\cite{Hungarian_algorithm} for alignment with ground-truth centers. As shown in~\Cref{tab:distance_of_predict_centers}, our approach achieves the lowest mean Hamming distance across all datasets (FFHQ $\to$ Pubfig83, FFHQ as $\mathcal{D}_{aux}$ and Pubfig83 as $\mathcal{D}_{priv}$) and code lengths. The improvement is most notable at 128 bits, where the distance is reduced by 10–30\% compared to K-means and by over 80\% compared to random assignment. These results highlight our method’s capacity to capture fine-grained semantic structures, whereas random centers fail to recover meaningful information. Supported by the distributions in~\cref{fig:distribution}, our findings confirm that the estimated centers offer accurate semantic guidance for inversion attacks, underscoring considerable privacy risks in deep hashing systems.

\subsubsection{Comparison of Model inversion Attack Methods}
We evaluate all methods using 64-bit hash codes across four dataset configurations, employing two evaluation models: FaceNet (tested on CelebA, PubFig83, and FFHQ→PubFig83) and VGG16 (on FaceScrub). As summarized in~\cref{tab:acc_of_predict_centers}, our approach consistently achieves superior performance. In FaceNet-based evaluations, it attains the highest Top-1 attack accuracy (79.6\%, 73.6\%, 56.0\%) and the lowest KNN distances. On FaceScrub with VGG16, it matches the best Top-5 accuracy (95.0\%) while remaining competitive across other metrics.

\begin{figure*}[htbp]
  \centering
  \includegraphics[width=0.9\linewidth]{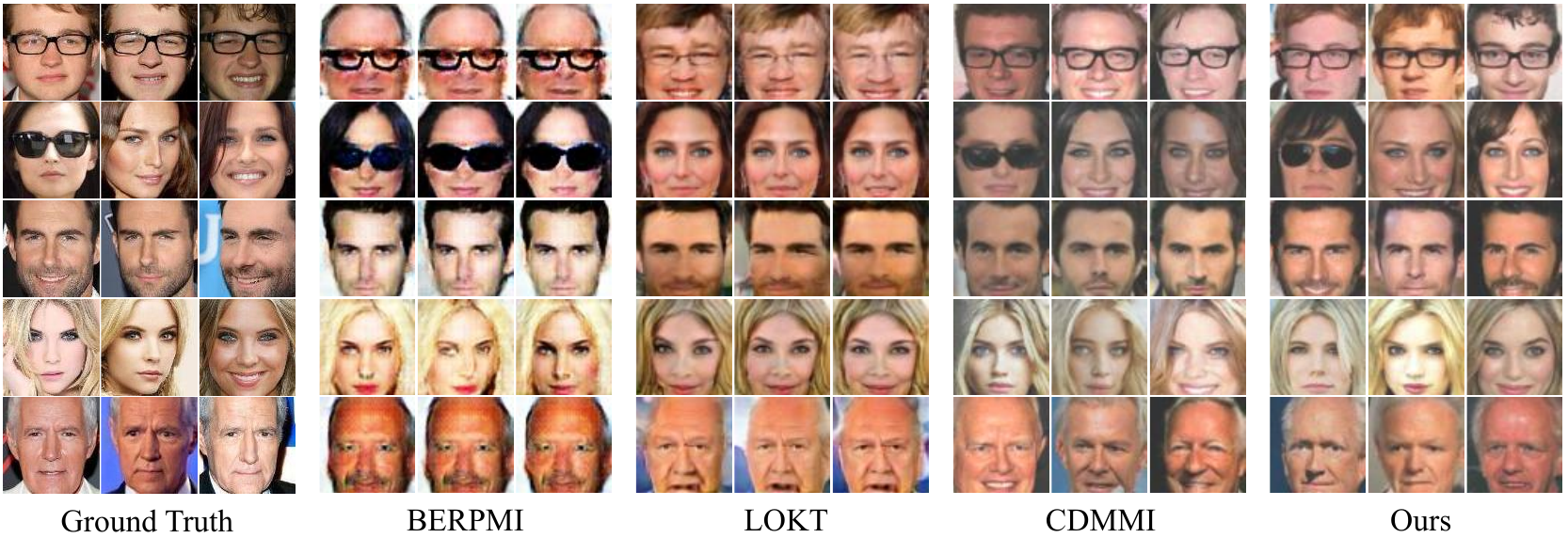}
  \caption{Comparison of Reconstructed Images}
   \label{fig:reconstructed_images}
\end{figure*}

As shown in~\cref{fig:reconstructed_images}, the images reconstructed by our method exhibit greater diversity than those produced by BERPMI and LOKT. This diversity enables a more accurate reflection of the true training data distribution, thereby revealing more privacy-sensitive information to attackers.

Overall, these results collectively demonstrate that our method effectively captures fundamental semantic features that generalize well across different datasets, thereby clearly confirming its robustness and broad applicability.


\subsection{Parameter Sensitivity Analysis}
\label{subsec:parameter_sensitivity}
We analyze the impact of hash code length $l$, a critical parameter influencing attack performance. As shown in~\cref{tab:length_analysis}, our method consistently outperforms all baseline approaches across various code lengths on the CelebA dataset. It achieves the highest Top-1 attack accuracy of 79.6\% at 64 bits, indicating an optimal balance between discriminative ability and reconstruction feasibility.

The performance gap is most pronounced at intermediate code lengths (32 and 64 bits), highlighting our method’s improved information utilization. Although all methods degrade at 128 bits due to increased sparsity in the Hamming space, our approach maintains the strongest results (72.2\%), demonstrating better adaptation to high-dimensional discrete optimization.

\begin{table}[htbp]
  \centering
  \caption{The Effect of Hash Length on Attack Accuracy (CelebA)}
  \label{tab:length_analysis}
  \begin{tabular}{lcccc}
    \toprule {Method}& 16bits & 32bits & 64bits & 128bits \\
    \midrule
    BERPMI    & 20.20\% & 36.00\% & 40.60\% & 36.20\% \\
    LOKT & 34.40\% & 61.80\% & 70.00\% & 70.60\% \\
    CDMMI  & 32.80\% & 44.20\% & 51.00\% & 50.80\% \\
    Ours  & \textbf{40.00}\% & \textbf{64.40}\% & \textbf{79.60}\% & \textbf{72.20}\% \\
    \bottomrule
  \end{tabular}
\end{table}


\subsection{Ablation Studies}
\label{subsec:ablation_studies}

We conduct ablation studies to validate two key components of DHMI: hash center selection and denoising optimization.

\textbf{Hash Center Selection:} Table~\ref{tab:ablation_center_strategies} compares different centers selection strategies on CelebA (64bits). Our slice-fused approach achieves 79.60\% with DHMI, significantly outperforming random selection (22.80\%) and K-means (56.80\%). This demonstrates our method's superior ability to capture semantic structure in Hamming space.

\textbf{Denoising Optimization:} Table~\ref{tab:ablatie_slice_fused} shows ablation results on CelebA (128-bit) with our slice-fused centers, comparing: baseline (CDMMI); an ablated variant without $S_{attack}$ (absence of $w_{base}$,$w_{haming}$); and our full method. Our approach outperforms across all metrics, achieving a 21.4\% absolute gain in Top-1 accuracy over the original baseline, validating our optimization strategy. The improved mAP further confirms our method reconstructs images that maintain high retrieval precision for deep hashing systems.

Both components are crucial: precise centers ensure semantic relevance, while guided optimization enables effective navigation of the non-differentiable Hamming space.

\begin{table}[htbp]
  \centering
  \caption{Ablation Study on Center Selection (CelebA, 64bits).}
  \label{tab:ablation_center_strategies}
  \begin{tabular}{lcccc}
    \toprule
    \multirow{2}{*}{Center} & \multicolumn{4}{c}{Method} \\
    \cmidrule(lr){2-5}
    & BERPMI & LOKT & CDMMI & DHMI\\
    \midrule
    Random    & 5.00\% & 27.80\% & 16.60\% & 22.80\% \\
    K-means & 23.00\% & 51.40\% & 50.20\% & 56.80\% \\
    Ours  & \textbf{40.60\%} & \textbf{70.00\%} & \textbf{51.00\%} & \textbf{79.60\%} \\
    \bottomrule
  \end{tabular}
\end{table}

\begin{table}[htbp]
  \centering
    \small 
  \setlength{\tabcolsep}{5pt} 
  \caption{Ablation Study on Denoising Optimization (CelebA)}
  \label{tab:ablatie_slice_fused}
  \begin{tabular}{lcccc}
    \toprule {Method}& Top-1 $\uparrow$ & Top-5  $\uparrow$ & KNN dist.  $\downarrow$ & mAP  $\uparrow$\\
    \midrule
    baseline    & 50.80\% & 75.20\% & 1266.20 & 76.93\%\\
    no $S_{attack}$  & 66.40\% & 83.00\% & 1184.24 & 86.23\%\\
    Ours  & \textbf{72.20}\% & \textbf{83.60}\% & \textbf{1159.36} & \textbf{86.77}\%\\
    \bottomrule
  \end{tabular}
\end{table}


\section{Conclusion}
\label{sec:conclusion}

In this paper, we propose DHMI, the first diffusion-based framework for model inversion attacks on deep hashing systems. DHMI first estimates semantic hash centers from an auxiliary dataset as surrogate anchors. It then employs a surrogate-guided denoising process, driven by a novel metric fusing classification consistency and hash proximity, to generate high-fidelity images. Extensive experiments demonstrate that DHMI effectively reconstructs high-resolution, privacy-sensitive images under a strict black-box setting, outperforming existing attacks and revealing critical privacy risks in deep hashing models. Future work will focus on improving the attack's generality and efficiency.

\bibliographystyle{IEEEtran}
\bibliography{reference}

@String(AAAI = {AAAI})

@InProceedings{Xia2014_2,
  author    = {Rongkai Xia and Yan Pan and Hanjiang Lai and Cong Liu and Shuicheng Yan},
  title     = {Supervised Hashing for Image Retrieval via Image Representation Learning},
  booktitle = {Proceedings of the Twenty-Eighth AAAI Conference on Artificial Intelligence},
  year      = 2014,
  volume    = 28,
  number    = 1,
  pages     = {2156-2162},
  doi       = {10.1609/aaai.v28i1.8952}
}

@InProceedings{Liu2016_3,
  author    = {Haomiao Liu and Ruiping Wang and Shiguang Shan and Xilin Chen},
  title     = {Deep supervised hashing for fast image retrieval},
  booktitle = {Proceedings of the 2016 IEEE Conference on Computer Vision and Pattern Recognition},
  year      = 2016,
  volume    = {},
  number    = {},
  pages     = {2064-2072},
  doi       = {10.1109/CVPR.2016.227}
}

@Article{Chen2025_4,
  author    = {Yinqi Chen and Zhiyi Lu and Yangting Zheng and Peiwen Li and Weijian Luo and Shuo Kang},
  title     = {Deep hashing with mutual information: A comprehensive strategy for image retrieval},
  journal   = {Expert Systems with Applications},
  year      = 2025,
  volume    = 264,
  number    = {},
  pages     = {125880},
  doi       = {10.1016/j.eswa.2024.125880}
}

@Article{Zhu2017_5,
  author    = {Zhu, Fuqing and Kong, Xiangwei and Zheng, Liang and Fu, Haiyan and Tian, Qi},
  journal   = {IEEE Transactions on Image Processing}, 
  title     = {Part-Based Deep Hashing for Large-Scale Person Re-Identification}, 
  year      = {2017},
  volume    = {26},
  number    = {10},
  pages     = {4806-4817},
  doi       = {10.1109/TIP.2017.2695101}
}

@InProceedings{Chen2017_6,
  author    = { Chen, Jiaxin and Wang, Yunhong and Qin, Jie and Liu, Li and Shao, Ling },
  title     = {Fast person re-identification via cross-camera semantic binary transformation},
  booktitle = {Proceedings of the 2017 IEEE Conference on Computer Vision and Pattern Recognition},
  year      = 2017,
  volume    = {},
  number    = {},
  pages     = {5330-5339},
  doi       = {10.1109/CVPR.2017.566}
}

@Article{Chen2024_7,
  author    = {Yuping Chen and Zhian He and Muhammad Awais Ashraf and Xinwen Chen and Yu Liu and Xiangting Ding and Binbin Tong and Yijie Chen},
  title     = {Performance evaluation of attention-deep hashing based medical image retrieval in brain MRI datasets},
  journal   = {Journal of Radiation Research and Applied Sciences},
  volume    = {17},
  number    = {3},
  pages     = {100968},
  year      = {2024},
  doi       = {10.1016/j.jrras.2024.100968},
}

@Article{Ghasemi2024_8,
  author    = {Mahsa Ghasemi and Hamid Hassanpour},
  title     = {FRIH: A face recognition framework using image hashing},
  journal   = {Multimedia Tools and Applications},
  year      = {2024},
  volume    = {83},
  number    = {21},
  pages     = {60147-60169},
  doi       = {10.1007/s11042-023-18007-9},
}

@Article{Borra2024_9,
  title     = {Deep hashing with multilayer CNN-based biometric authentication for identifying individuals in transportation security},
  author    = {Borra, Subba Reddy and Premalatha, B and Divya, G and Srinivasarao, B and Eshwar, D and Reddy, V Bharath Simha and Kumar, Pala Mahesh},
  journal   = {Journal of Transportation Security},
  volume    = {17},
  number    = {1},
  pages     = {4},
  year      = {2024},
  doi       = {10.1007/s12198-024-00272-w}
}

@InProceedings{Fredrikson2014_10,
  author    = {Matthew Fredrikson and Eric Lantz and Somesh Jha and Simon Lin and David Page and Thomas Ristenpart},
  title     = {Privacy in Pharmacogenetics: An End-to-End Case Study of Personalized Warfarin Dosing},
  booktitle = {Proceedings of the 23rd USENIX Conference on Security Symposium},
  year      = {2014},
  pages     = {17-32},
  doi       = {10.5555/2671225.2671227}
}

@InProceedings{Fredrikson2015_11,
  author    = {Fredrikson, Matt and Jha, Somesh and Ristenpart, Thomas},
  title     = {Model Inversion Attacks that Exploit Confidence Information and Basic Countermeasures},
  booktitle = {Proceedings of the 22nd ACM SIGSAC Conference on Computer and Communications Security},
  year      = {2015},
  pages     = {1322–1333},
  doi       = {10.1145/2810103.2813677}
}

@Article{Subbanna2021_13,
  author    = {Subbanna, Nagesh and Wilms, Matthias and Tuladhar, Anup and Forkert, Nils D.},
  title     = {An Analysis of the Vulnerability of Two Common Deep Learning-Based Medical Image Segmentation Techniques to Model Inversion Attacks},
  journal   = {Sensors},
  volume    = {21},
  year      = {2021},
  number    = {11},
  doi       = {10.3390/s21113874}
}

@InProceedings{Zhang2020_14,
  author    = {Zhang, Yuheng and Jia, Ruoxi and Pei, Hengzhi and Wang, Wenxiao and Li, Bo and Song, Dawn},
  booktitle = {Proceedings of the 2020 IEEE/CVF Conference on Computer Vision and Pattern Recognition}, 
  title     = {The Secret Revealer: Generative Model-Inversion Attacks Against Deep Neural Networks}, 
  year      = {2020},
  volume    = {},
  number    = {},
  pages     = {250-258},
  doi={10.1109/CVPR42600.2020.00033}
}

@Article{Liu2024_15,
  author    = {Liu, Rongke and Wang, Dong and Ren, Yizhi and Wang, Zhen and Guo, Kaitian and Qin, Qianqian and Liu, Xiaolei},
  journal   = {IEEE Transactions on Information Forensics and Security}, 
  title     = {Unstoppable Attack: Label-Only Model Inversion Via Conditional Diffusion Model}, 
  year      = {2024},
  volume    = {19},
  number    = {},
  pages     = {3958-3973},
  doi={10.1109/TIFS.2024.3372815}
}

@InProceedings{Li2017_16,
  author    = {Li, Qi and Sun, Zhenan and He, Ran and Tan, Tieniu},
  title     = {Deep supervised discrete hashing},
  year      = {2017},
  booktitle = {Proceedings of the 31st International Conference on Neural Information Processing Systems},
  pages     = {2479–2488},
  doi       = {10.5555/3294996.3295009}
}

@InProceedings{Wang2017_17,
  author    = {Wang, Xiaofang and Shi, Yi and Kitani, Kris M.},
  title     = {Deep Supervised Hashing with Triplet Labels},
  booktitle = {Asian conference on computer vision},
  year      = {2017},
  pages     = {70-84},
  doi       = {10.1007/978-3-319-54181-5_5}
}

@InProceedings{Jain2017_18,
  author    = {Jain, Himalaya and Zepeda, Joaquin and Pérez, Patrick and Gribonval, Rémi},
  title     = {SuBiC: A Supervised, Structured Binary Code for Image Search}, 
  booktitle = {Proceedings of the 2017 IEEE International Conference on Computer Vision},
  year      = {2017},
  volume    = {},
  number    = {},
  pages     = {833-842},
  doi       = {10.1109/ICCV.2017.96}
}

@InProceedings{Yuan2020_19,
  author    = {Yuan, Li and Wang, Tao and Zhang, Xiaopeng and Tay, Francis EH and Jie, Zequn and Liu, Wei and Feng, Jiashi},
  booktitle = {Proceedings of the 2020 IEEE/CVF Conference on Computer Vision and Pattern Recognition}, 
  title     = {Central Similarity Quantization for Efficient Image and Video Retrieval}, 
  year      = {2020},
  volume    = {},
  number    = {},
  pages     = {3080-3089},
  doi       = {10.1109/CVPR42600.2020.00315}
}

@InProceedings{Wang2023_20,
  author    = {Wang, Liangdao and Pan, Yan and Liu, Cong and Lai, Hanjiang and Yin, Jian and Liu, Ye},
  booktitle = {Proceedings of the 2023 IEEE/CVF Conference on Computer Vision and Pattern Recognition}, 
  title     = {Deep Hashing with Minimal-Distance-Separated Hash Centers}, 
  year      = {2023},
  volume    = {},
  number    = {},
  pages     = {23455-23464},
  doi       = {10.1109/CVPR52729.2023.02246}
}

@InProceedings{Klein2019_21,
  author    = {Klein, Benjamin and Wolf, Lior},
  booktitle = {Proceedings of the 2019 IEEE/CVF Conference on Computer Vision and Pattern Recognition}, 
  title     = {End-To-End Supervised Product Quantization for Image Search and Retrieval}, 
  year      = {2019},
  volume    = {},
  number    = {},
  pages     = {5036-5045},
  doi       = {10.1109/CVPR.2019.00518}
}

@InProceedings{Yang2018_22,
  author    = {Yang, Erkun and Deng, Cheng and Liu, Tongliang and Liu, Wei and Tao, Dacheng},
  title     = {Semantic structure-based unsupervised deep hashing},
  year      = {2018},
  booktitle = {Proceedings of the 27th International Joint Conference on Artificial Intelligence},
  pages     = {1064–1070},
  doi       = {10.5555/3304415.3304566}
}

@InProceedings{Hu2017_23,
  author    = {Hu, Qinghao and Wu, Jiaxiang and Cheng, Jian and Wu, Lifang and Lu, Hanqing},
  title     = {Pseudo Label based Unsupervised Deep Discriminative Hashing for Image Retrieval},
  year      = {2017},
  booktitle = {Proceedings of the 25th ACM International Conference on Multimedia},
  pages     = {1584–1590},
  doi       = {10.1145/3123266.3123403}
}

@InProceedings{Dizaji2018_24,
  author    = {Dizaji, Kamran Ghasedi and Zheng, Feng and Nourabadi, Najmeh Sadoughi and Yang, Yanhua and Deng, Cheng and Huang, Heng},
  booktitle = {Proceedings of the 2018 IEEE/CVF Conference on Computer Vision and Pattern Recognition}, 
  title     = {Unsupervised Deep Generative Adversarial Hashing Network}, 
  year      = {2018},
  volume    = {},
  number    = {},
  pages     = {3664-3673},
  doi={10.1109/CVPR.2018.00386}
}

@InProceedings{Wang2021_25,
  author    = {Wang, Kuan-Chieh and Fu, Yan and Li, Ke and Khisti, Ashish and Zemel, Richard and Makhzani, Alireza},
  title     = {Variational model inversion attacks},
  booktitle = {Proceedings of the 35th International Conference on Neural Information Processing Systems},
  year      = {2021},
  volume    = {},
  number    = {},
  pages     = {9706-9719},
  doi       = {10.5555/3540261.3541004}
}

@InProceedings{Chen2021_26,
  author    = {Chen, Si and Kahla, Mostafa and Jia, Ruoxi and Qi, Guo-Jun},
  booktitle = {Proceedings of the 2021 IEEE/CVF International Conference on Computer Vision}, 
  title     = {Knowledge-Enriched Distributional Model Inversion Attacks}, 
  year      = {2021},
  volume    = {},
  number    = {},
  pages     = {16158-16167},
  doi       = {10.1109/ICCV48922.2021.01587}
}

@InProceedings{Yuan2023_27, 
  author    = {Yuan, Xiaojian and Chen, Kejiang and Zhang, Jie and Zhang, Weiming and Yu, Nenghai and Zhang, Yang}, 
  title     = {Pseudo Label-Guided Model Inversion Attack via Conditional Generative Adversarial Network}, 
  booktitle = {Proceedings of the AAAI Conference on Artificial Intelligence}, 
  year      = {2023},
  volume    = {37}, 
  number    = {3}, 
  pages     = {3349-3357},
  doi       = {10.1609/aaai.v37i3.25442}
}

@InProceedings{Yang2019_28,
  author    = {Yang, Ziqi and Zhang, Jiyi and Chang, Ee-Chien and Liang, Zhenkai},
  title     = {Neural Network Inversion in Adversarial Setting via Background Knowledge Alignment},
  booktitle = {Proceedings of the 2019 ACM SIGSAC Conference on Computer and Communications Security},
  year      = {2019},
  pages     = {225–240},
  doi       = {10.1145/3319535.3354261},
}

@InProceedings{Yoshimura2021_29,
  author    = {Yoshimura, Shunsuke and Nakamura, Kazuaki and Nitta, Naoko and Babaguchi, Noboru},
  booktitle = {Proceedings of the 2021 Asia-Pacific Signal and Information Processing Association Annual Summit and Conference}, 
  title     = {Model Inversion Attack against a Face Recognition System in a Black-Box Setting}, 
  year      = {2021},
  volume    = {},
  number    = {},
  pages     = {1800-1807},
  doi       = {}
}

@Article{dionysiou2023_30,
  author    = {Dionysiou, Antreas and Vassiliades, Vassilis and Athanasopoulos, Elias},
  title     = {Exploring model inversion attacks in the black-box setting},
  journal   = {Proceedings on Privacy Enhancing Technologies},
  year      = {2023},
  volume    = {2023},
  number    = {1},
  pages     = {190–206},
  doi       = {10.56553/popets-2023-0012}
}

@Article{Bao2025_31,
  author    = {Bao, Huan and Wei, Kaimin and Wu, Yongdong and Qian, Jin and Deng, Robert H.},
  journal   = {IEEE Transactions on Information Forensics and Security}, 
  title     = {Distributional Black-Box Model Inversion Attack With Multi-Agent Reinforcement Learning}, 
  year      = {2025},
  volume    = {20},
  number    = {},
  pages     = {5425-5437},
  doi={10.1109/TIFS.2025.3564043}
}

@InProceedings{Kahla2022_32,
  author    = {Kahla, Mostafa and Chen, Si and Just, Hoang Anh and Jia, Ruoxi},
  booktitle = {Proceedings of the 2022 IEEE/CVF Conference on Computer Vision and Pattern Recognition},
  title     = {Label-Only Model Inversion Attacks via Boundary Repulsion},
  year      = {2022},
  volume    = {},
  pages     = {15025-15033},
  doi       = {10.1109/CVPR52688.2022.01462},
}

@InProceedings{Nguyen2023_33,
  author    = {Nguyen, Ngoc-Bao and Chandrasegaran, Keshigeyan and Abdollahzadeh, Milad and Cheung, Ngai-Man},
  title     = {Label-only model inversion attacks via knowledge transfer},
  booktitle = {Proceedings of the 37th International Conference on Neural Information Processing Systems},
  year      = {2023},
  pages     = {68895-68907},
  doi       = {10.5555/3666122.3669137}
}

@InProceedings{Celeba,
 author = {Liu, Ziwei and Luo, Ping and Wang, Xiaogang and Tang, Xiaoou},
 title = {Deep Learning Face Attributes in the Wild},
 booktitle = {Proceedings of International Conference on Computer Vision},
 year = {2015},
 pages = {3730-3738},
 doi = {}
}

@InProceedings{FaceScrub,
  author={Ng, Hong-Wei and Winkler, Stefan},
  title={A data-driven approach to cleaning large face datasets}, 
  booktitle={2014 IEEE International Conference on Image Processing}, 
  year={2014},
  volume={},
  number={},
  pages={343-347},
  doi={10.1109/ICIP.2014.7025068}
}

@InProceedings{Pubfig83,
  author={Pinto, Nicolas and Stone, Zak and Zickler, Todd and Cox, David},
  title={Scaling up biologically-inspired computer vision: A case study in unconstrained face recognition on facebook}, 
  booktitle={2011 IEEE Computer Society Conference on Computer Vision and Pattern Recognition Workshops}, 
  year={2011},
  volume={},
  number={},
  pages={35-42},
  doi={10.1109/CVPRW.2011.5981788}
}

@Article{backdoor_deephash,
  author={Zhou, Ziqi and Deng, Menghao and Song, Yufei and Zhang, Hangtao and Wan, Wei and Hu, Shengshan and Li, Minghui and Yu Zhang, Leo and Yao, Dezhong},
  journal={IEEE Transactions on Information Forensics and Security}, 
  title={DarkHash: A Data-Free Backdoor Attack Against Deep Hashing}, 
  year={2025},
  volume={20},
  number={},
  pages={8139-8153}
}

@Article{gen_collision_attack_deephash,
  author={Ying, Luyang and Xiong, Cheng and Qin, Chuan and Luo, Xiangyang and Qian, Zhenxing and Zhang, Xinpeng},
  journal={IEEE Transactions on Information Forensics and Security}, 
  title={Generative Collision Attack on Deep Image Hashing}, 
  year={2025},
  volume={20},
  number={},
  pages={2748-2762}
}

@InProceedings{ffhq,
author = {Karras, Tero and Laine, Samuli and Aila, Timo},
title = {A Style-Based Generator Architecture for Generative Adversarial Networks},
booktitle = {Proceedings of the IEEE/CVF Conference on Computer Vision and Pattern Recognition},
month = {June},
year = {2019},
volume={},
number={},
pages={4401-4410},
}

@InProceedings{Resnet,
author = {He, Kaiming and Zhang, Xiangyu and Ren, Shaoqing and Sun, Jian},
title = {Deep Residual Learning for Image Recognition},
booktitle = {Proceedings of the IEEE Conference on Computer Vision and Pattern Recognition},
month = {June},
year = {2016},
pages={770-778},
}

@InProceedings{CDM,
title={Classifier-Free Diffusion Guidance},
author={Jonathan Ho and Tim Salimans},
booktitle={NeurIPS 2021 Workshop on Deep Generative Models and Downstream Applications},
year={2021}
}

@InProceedings{Densenet,
author = {Huang, Gao and Liu, Zhuang and van der Maaten, Laurens and Weinberger, Kilian Q.},
title = {Densely Connected Convolutional Networks},
booktitle = {Proceedings of the IEEE Conference on Computer Vision and Pattern Recognition},
month = {July},
year = {2017},
pages = {4700-4708}
}

@InProceedings{FaceNet,
author = {Schroff, Florian and Kalenichenko, Dmitry and Philbin, James},
title = {FaceNet: A Unified Embedding for Face Recognition and Clustering},
booktitle = {Proceedings of the IEEE Conference on Computer Vision and Pattern Recognition},
month = {June},
year = {2015},
pages = {815-823}
}

@InProceedings{VGG16,
  title = {Very deep convolutional networks for large-scale image recognition},
  booktitle = {3rd International Conference on Learning Representations},
  volume = {},
  author = {Simonyan, K and Zisserman, A},
  year = {2015},
  pages = {1-14}
}

@article{Hungarian_algorithm,
author = {Kuhn, H. W.},
title = {The Hungarian method for the assignment problem},
journal = {Naval Research Logistics Quarterly},
volume = {2},
number = {1-2},
pages = {83-97},
year = {1955}
}

@InProceedings{kmeans,
  title={Some methods of classification and analysis of multivariate observations},
  booktitle={Proc. of 5th Berkeley Symposium on Math. Stat. and Prob.},
  author={James MacQueen},
  pages={281--297},
  year={1967}
}

\clearpage
\setcounter{page}{1}

\section{Additional Experiment Details}
\label{sec:details}

\subsection{Dataset Setup}
\label{subsec:dataset}

The dataset configuration in this paper primarily follows \cite{Nguyen2023_33}. Specifically, for \textbf{CelebA}, we employ the \textbf{first 100 classes} as the private dataset $\mathcal{D}_{priv}$ to balance the retrieval precision of the target deep hashing model. For \textbf{FaceScrub}, we utilize the available 43,124 images, designating its 200 classes as $\mathcal{D}_{priv}$ and the remaining 330 classes as $\mathcal{D}_{aux}$. In \textbf{PubFig83}, the 50 classes are used as $\mathcal{D}_{priv}$, while the remaining 33 classes serve as $\mathcal{D}_{aux}$. Additionally, the first 10,000 images from the \textbf{FFHQ} dataset are incorporated into $\mathcal{D}_{aux}$. The detailed statistics for these splits are summarized in \textbf{Table~\ref{tab:dataset_stats}}.

\begin{table}[htbp]
\centering
\caption{Dataset Splits Used in the Experiments.}
\label{tab:dataset_stats}
\begin{tabular}{lcccc}
\toprule
Dataset & \multicolumn{1}{c}{Classes} & \multicolumn{1}{c}{\({\mathcal{D}_{priv}^{train}}\)} & \multicolumn{1}{c}{\({\mathcal{D}_{priv}^{test}}\)} & \multicolumn{1}{c}{\(\mathcal{D}_{{aux}}\)} \\
\midrule
CelebA & 100 & 2,700 & 300 & 30,000 \\
FaceScrub & 200 & 11,275 & 4,973 & 26,876 \\
PubFig83 & 50 & 6,482 & 1,642 & 5,679 \\
\bottomrule
\end{tabular}
\end{table}


\subsection{Evaluation Model Setup}
\label{subsec:eval_model}

The evaluation models and their configurations in this paper are established as in \cite{Nguyen2023_33}. For the \textbf{CelebA} dataset, we employ \textbf{FaceNet} as the evaluator due to its high classification accuracy (\textbf{97.27\%}), ensuring precise facial identity recognition. The backbone for its deep hashing-based retrieval task is \textbf{ResNet-101}, which yields a retrieval precision of \textbf{84.76\%}. Similarly, for both the \textbf{PubFig83} and the \textbf{FFHQ $\to$ PubFig83} evaluations, \textbf{FaceNet} is adopted (achieving \textbf{95.81\%} accuracy) to maintain a consistent and highly accurate evaluation standard. In contrast, the \textbf{FaceScrub} dataset is evaluated using a \textbf{VGG-16} model (with \textbf{88.32\%} accuracy) to investigate the impact of different evaluation model architectures on the experimental outcomes. It is important to note that all evaluation models are trained on the same training set as the target deep hashing model, ensuring a fair and consistent comparison. Regarding the input sizes, FaceNet models operate on $112\times112$ images, while all other models use $64\times64$ inputs.

\subsection{Attack Model Setup}
\label{subsec:attack_model}

The implementation details of the conditional diffusion model follow the configuration in \cite{Liu2024_15}. The model employs a U-Net backbone, trained for 300 iterations with an initial learning rate of $3\times10^{-4}$ that is reduced to $1\times10^{-4}$ for the final 50 iterations. Optimization is performed using the MSE loss function and the AdamW optimizer with an Exponential Moving Average mechanism. The diffusion process is configured with $T=1500$ noise steps and a linear variance schedule where $\beta_0 = 1\times10^{-4}$ and $\beta_t = 0.02$. Additionally, we use a guidance strength $\omega=4$ and a probability parameter $p=0.1$.

For the GAN and the cluster of surrogate models, our configurations adhere to \cite{Nguyen2023_33}. The generated dataset for training the surrogate models comprises 500,000 images.


\section{Additional Analysis and Visualizations}
\label{sec:additional_results}

\subsection{Cross-Model Analysis}
\label{subsec:cross_model_analysis}

\begin{table}[htbp]
  \centering
  \caption{Mean Hamming Distance on Different Target Models}
  \label{tab:differnet_models_hamming_dist}
  \begin{tabular}{lccc}
    \toprule {Model}& Random & K-means & Ours\\
    \midrule
    ResNet-50    & 22.75 & 2.90 & \textbf{2.41} \\
    VGG-16    & 22.54 & 4.43 & \textbf{3.33} \\
    EfficientNet-B0    & 22.66 & 4.79 & \textbf{3.94}\\
    \bottomrule
  \end{tabular}
\end{table}

\begin{table}[htbp]
  \centering
    \small 
  \setlength{\tabcolsep}{5pt} 
  \caption{The Effect of Different Target Models}
  \label{tab:different_target_models}
  \begin{tabular}{lcccc}
    \toprule {Model}& Top-1 $\uparrow$ & Top-5  $\uparrow$ & KNN dist.  $\downarrow$ & mAP  $\uparrow$\\
    \midrule
    ResNet-50    & 79.60\% & 88.40\% & 1086.26 & 88.00\%\\
    VGG-16  & 61.00\% & 81.20\% & 1261.23 & 81.36\%\\
    EfficientNet-B0  & 55.80\% & 78.20\% & 1289.94 & 78.64\%\\
    \bottomrule
  \end{tabular}
\end{table}

\begin{figure*}[htbp]
  \centering
  \includegraphics[width=1\linewidth]{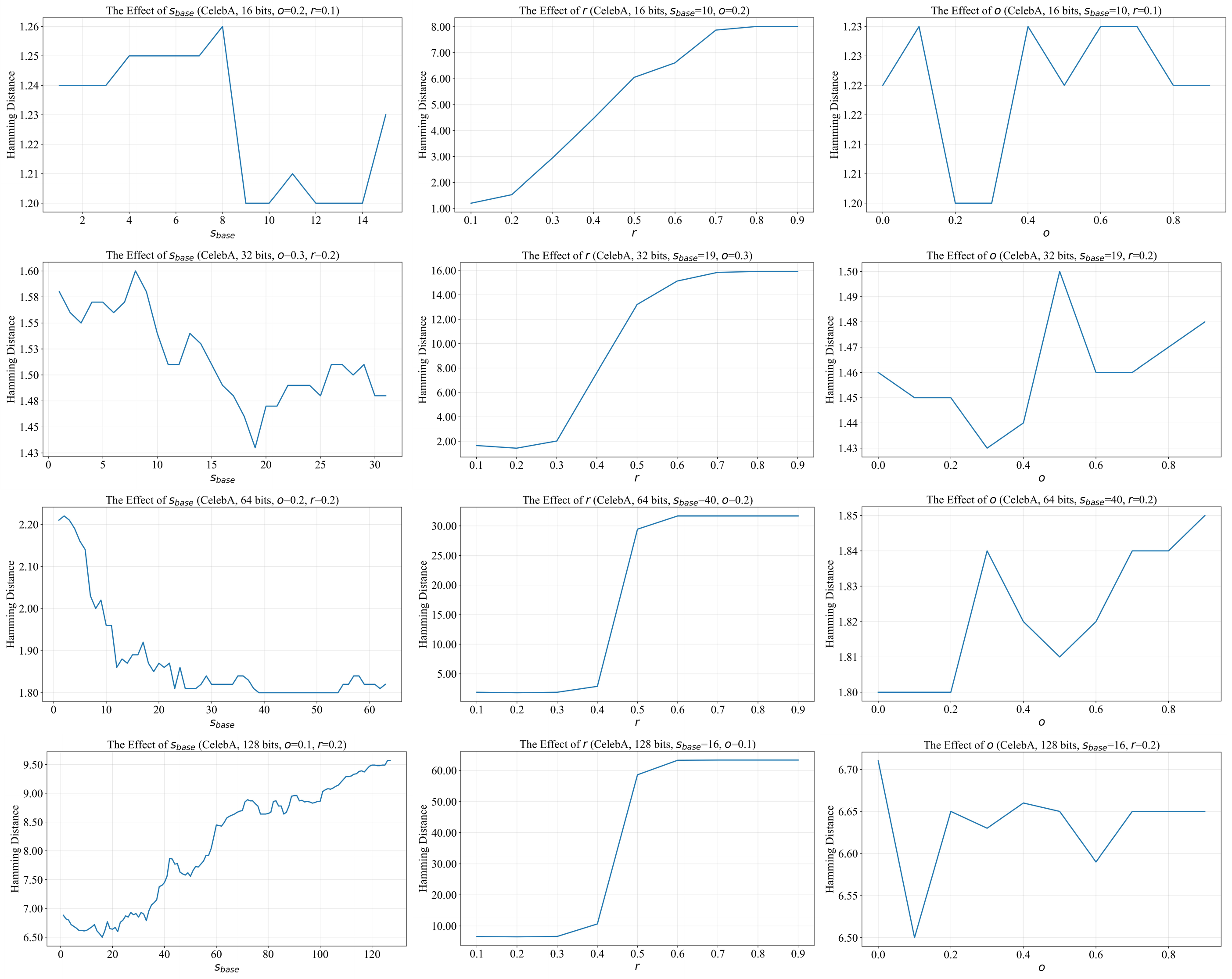}
  \caption{Analysis of Parameter Effects in the Slice-Fused Hash Center Estimation Method}
   \label{fig:different_parameters}
\end{figure*}

We further evaluate different backbone architectures for the target deep hashing model on the \textbf{CelebA} dataset with \textbf{64-bit} hash codes, following the experimental setup in~\cite{Wang2023_20}. The mean Average Precision (mAP@all) values for \textbf{ResNet-50}, \textbf{VGG-16}, and \textbf{EfficientNet-B0} are \textbf{83.91\%}, \textbf{83.61\%}, and \textbf{84.12\%}, respectively. The corresponding average Hamming distances between the predicted hash centers and the true centers are summarized in \textbf{Table~\ref{tab:differnet_models_hamming_dist}}, while the overall inversion performance metrics across different model architectures are presented in \textbf{Table~\ref{tab:different_target_models}}.

The performance variations in Table~\ref{tab:different_target_models}, where \textbf{ResNet-50} outperforms \textbf{VGG-16} and \textbf{EfficientNet-B0}, are directly attributable to differences in hash center prediction quality. As shown by the Mean Hamming Distances in Table~\ref{tab:differnet_models_hamming_dist}, the centers predicted using VGG-16 and EfficientNet-B0 exhibit larger Hamming distances from the true centers than those from ResNet-50. This gap arises from their distinct architectural inductive biases: VGG-16's plain deep structure may lead to less robust feature learning, while EfficientNet-B0's parameter efficiency focus might compromise the feature richness required for precise center allocation.

Critically, although the accuracy of our hash center prediction is influenced by the model architecture, our method consistently and significantly outperforms the baseline approaches (Random initialization and K-means clustering) across all three backbones. This compelling evidence, demonstrated by the lower Hamming distances in Table~\ref{tab:differnet_models_hamming_dist} and the exceptional performance in Table~\ref{tab:different_target_models} across all backbones, confirms the fundamental robustness and general applicability of our proposed framework for both center estimation and image reconstruction.

\begin{figure*}[htbp]
  \centering
  \includegraphics[width=0.9\linewidth]{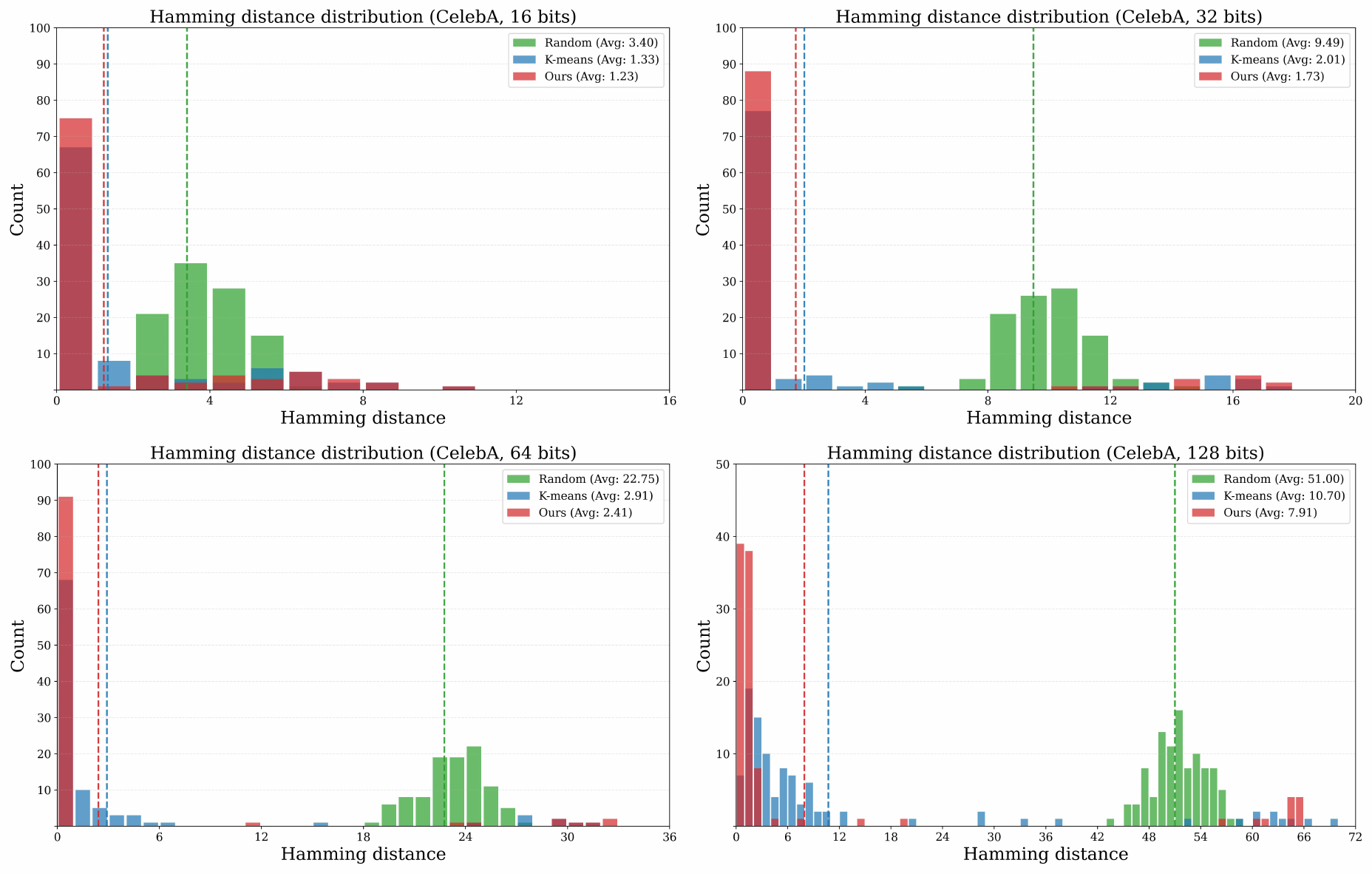}
  \caption{Analysis of Hamming Distance Distributions for Different Prediction Methods}
   \label{fig:hamming_distance_distribution}
\end{figure*}

\subsection{Parameter Analysis of Slice-Fused}
\label{subsec:parameter_analysis}

We first conducted a grid search experiment to identify the optimal combination of the three parameters ($s_{base}$, $r$, and $o$). To systematically investigate the individual impact of each parameter on the Hamming distance, we then fixed two parameters at their optimal values while varying the third. The following subsections detail the findings of this parameter sensitivity analysis.

\subsubsection{The Effect of Base Slice Size}

The parameter $s_{base}$, which controls the base slice size in our estimation method, demonstrates a non-monotonic relationship with the Hamming Distance across different bit lengths. As shown in the left column of Figure~\ref{fig:different_parameters}, the Hamming Distance initially decreases as $s_{base}$ increases, reaching an optimal point, after which it begins to rise again. This pattern indicates that there exists a specific $s_{base}$ value that minimizes the alignment error. When $s_{base}$ is too small, the estimation may be too sensitive to local variations, while an excessively large $s_{base}$ can obscure important fine-grained patterns. The optimal value $s_{base}$ tends to increase with higher bit lengths, suggesting that more complex representations benefit from analyzing larger slices of the hash code.

\subsubsection{The Effect of Neighborhood Radius Ratio}

The neighborhood radius ratio $r$ demonstrates a critical threshold behavior in its impact on Hamming Distance. As evidenced in the middle column of Figure~\ref{fig:different_parameters}, the Hamming Distance remains stable at a low level when $r$ is below a certain critical value. However, once $r$ exceeds this threshold, a sharp increase in Hamming Distance is observed. This phenomenon suggests that $r$ controls the sensitivity of the estimation process. Within the optimal range, $r$ enables the method to capture meaningful local patterns while filtering out noise. Beyond the critical point, the neighborhood becomes overly inclusive, incorporating irrelevant patterns that degrade estimation quality. Notably, higher bit lengths can tolerate larger $r$ values before performance degradation occurs, as the expanded representation space in longer codes requires a larger neighborhood radius to encompass a sufficient number of sample points for capturing meaningful local patterns.

\subsubsection{The Effect of Slice Overlap Ratio}

The overlap ratio $o$ exhibits a clear non-monotonic relationship with Hamming Distance across most bit lengths, as demonstrated in the right column of Figure~\ref{fig:different_parameters}. This relationship is characterized by an initial decrease in distance as $o$ increases, followed by a rise after exceeding a certain threshold. The 16-bit, 32-bit, and 128-bit configurations all display this pattern, where the Hamming Distance first declines to a distinct minimum before increasing again, indicating the presence of an optimal overlap value that minimizes alignment error. In contrast, the 64-bit configuration shows multi-modal fluctuations without a clear global minimum, suggesting higher optimization instability in this intermediate dimensionality. These observations highlight that the optimal $o$ is highly dependent on bit length, with higher-dimensional spaces requiring careful overlap tuning to balance pattern coverage and estimation robustness.

Our parameter analysis reveals distinct roles for the three parameters: $s_{base}$ controls pattern granularity with an optimal mid-range, $r$ acts as a critical threshold requiring careful bounding, and $o$ exhibits bit-length-dependent behavior needing a specific configuration. In the main experiments, we selected parameter values that prioritize robust performance across varied conditions rather than pursuing absolute optimums. This practical approach is particularly important in real-world black-box scenarios where exhaustive parameter tuning is infeasible, ensuring our method's applicability when perfect parameter optimization is not possible.


\subsection{Analysis of Hamming Distance Distributions}
\label{subsec:distribution_different_methods}

Figure~\ref{fig:hamming_distance_distribution} presents the Hamming distance distributions between predicted hash centers and ground-truth centers on the CelebA dataset across 16, 32, 64, and 128-bit configurations, comparing our proposed method with K-means-based prediction and random initialization.

The alignment performance shows a notable trend as the bit length increases from 16 to 64 bits. Our method achieves 75 exact matches (1.23 average distance) at 16 bits, improving to 88 matches (1.73 average distance) at 32 bits, and reaching optimal performance with 91 matches (2.41 average distance) at 64 bits. This improvement stems from the expanded representation capacity that allows for more precise semantic encoding as dimensionality increases.

However, at 128 bits, the performance declines to 39 matches (7.91 average distance), which can be attributed to the increased optimization complexity and the challenge of maintaining accurate semantic relationships in ultra-high-dimensional spaces. Despite this decline, our method maintains a substantial advantage over K-means (7 matches, 10.70 average distance) and random initialization.

The distribution patterns in Figure~\ref{fig:hamming_distance_distribution} visually corroborate these trends, showing concentrated low-distance distributions for our method across all bit lengths, with optimal concentration observed at 32-64 bits and moderate dispersion at 128 bits. These distribution characteristics provide strong support for the effectiveness of our method, demonstrating its superior capability in hash center estimation particularly in low-to-medium dimensional spaces while maintaining competitive performance even in challenging high-dimensional scenarios.


\subsection{Parameter Analysis of Surrogate-Driven}
\label{subsec:parameter_of_surrogate_driven}
The hyperparameters in our framework were configured through preliminary experiments to establish a well-performing configuration under computational constraints. To systematically evaluate parameter sensitivity, we conducted controlled experiments on the CelebA dataset with 64-bit hash codes using our slice-fused estimation method. All parameters except the one being analyzed were fixed at our established experimental configuration values, enabling isolation of individual parameter impacts.

To minimize external interference and enhance computational efficiency, we employed fixed random seeds throughout the denoising process and saved intermediate states. This ensures reproducible results by enabling direct computation from any intermediate step. \textbf{It should be noted that due to this methodological approach, metrics in this section may exhibit slight discrepancies from those in the main text.}

\subsubsection{The Effect of Optimize Times }

The optimization times $iter$ exhibit a significant impact on model performance, as detailed in Table~\ref{tab:iter}. The most substantial improvements occur in the first two iterations, with Top-1 attack accuracy increasing by 18.60\% and $S_{attack}$ by 0.1205, while KNN distance decreases by 95.37. Beyond this point, gains become more gradual: Top-1 attack accuracy peaks at 8 iterations (76.80\%), while Top-5 and mAP achieve optimal values at 2 iterations (89.80\% and 89.78\%).

Notably, our attack adaptation metric ($S_{attack}$) shows continuous improvement, reaching 0.6117 at 10 iterations. However, since other critical metrics have stabilized and show no significant further improvement beyond 6-8 iterations, additional optimization yields diminishing returns for overall performance. The KNN distance stabilizes around 1082-1086 after initial optimization. Based on these trends, 6-8 iterations provide the optimal balance, achieving competitive $S_{attack}$ performance while maintaining efficiency and avoiding unnecessary computation when other metrics have converged.

\begin{table}[htbp]
  \centering
    \small 
  \setlength{\tabcolsep}{4.5pt} 
  \caption{The Effect of Optimize Times}
  \label{tab:iter}
  \begin{tabular}{lccccc}
    \toprule {$iter$}& Top-1 $\uparrow$ & Top-5  $\uparrow$ & KNN dist.  $\downarrow$ & mAP  $\uparrow$ & $S_{attack}$  $\uparrow$\\
    \midrule
    0    & 55.20\% & 80.40\% & 1178.23 & 78.08\% & 0.4572\\
    2    & 73.80\% & \textbf{89.80}\% & 1082.86 & \textbf{89.78}\% & 0.5777\\
    4  & 74.60\% & 89.40\% & 1085.95 & 89.27\% & 0.5954\\
    6  & 76.20\% & 88.60\% & 1082.25 & 89.54\% & 0.6066\\
    8  & \textbf{76.80}\% & 89.00\% & 1084.95 & 88.98\% & 0.6081\\
    10  & 76.00\% & 88.80\% & \textbf{1081.82} & 88.95\% & \textbf{0.6117}\\
    \bottomrule
  \end{tabular}
\end{table}


\subsubsection{The Effect of Surrogate Model Cluster Size}
As shown in Table~\ref{tab:m}, the surrogate model cluster size $m$ demonstrates a clear performance trend. Top-1 attack accuracy improves consistently from 71.20\% at $m=1$ to a peak of 76.20\% at $m=3$, after which it stabilizes at 75.80\%. Notably, our attack adaptation metric ($S_{attack}$) reaches its maximum of 0.6066 at $m=3$. In contrast, Top-5 attack accuracy shows more gradual improvement, achieving its best performance of 89.40\% at $m=4$.

These results indicate that increasing the number of surrogate models enhances performance up to a point, with $m=3$ representing the optimal balance between model diversity and computational efficiency. The performance saturation beyond $m=3$ suggests diminishing returns from additional models, possibly due to increased noise or redundancy in the cluster predictions.

\begin{table}[htbp]
  \centering
    \small 
  \setlength{\tabcolsep}{4pt} 
  \caption{The Effect of Surrogate Model Cluster Size}
  \label{tab:m}
  \begin{tabular}{lccccc}
    \toprule {Metric}& $m=1$ & $m=2$ & $m=3$ & $m=4$ & $m=5$\\
    \midrule
    Top-1    & 71.20\% & 73.60\% & \textbf{76.20}\% & 75.80\%  & 75.80\%\\
    Top-5   & 88.00\% & 88.00\% & 88.60\% & \textbf{89.40}\%  & 89.20\%\\
    $S_{attack}$ & 0.5723 & 0.5896 & \textbf{0.6066} & 0.5974 & 0.5980 \\
    \bottomrule
  \end{tabular}
\end{table}


\subsubsection{The Effect of Base Prediction Weight}
Table~\ref{tab:w_base} evaluates the impact of the base prediction weight \(w_{base}\) by comparing the performance achieved using only the selection operation at step 12 in~\Cref{alg:denoise} (\textbf{Before}) with that after incorporating the surrogate models optimization (\textbf{After}). The parameter \(w_{base}\) is introduced to balance the initial prediction from the selection step with the refined output from the surrogate models optimization. As $w_{base}$ increases from 0.0 to 0.4, the performance under the selection-only configuration improves consistently, reaching optimal Top-1 (55.40\%) and Top-5 (80.40\%) attack accuracy at \(w_{base} \geq 0.3\).

The surrogate models optimization substantially enhances Top-1 attack accuracy across all weights, peaking at 76.40\% for \(w_{base} \geq 0.3\). In contrast, Top-5 attack accuracy achieves its maximum (89.40\%) at lower weights (\(w_{base} \leq 0.1\)). These results validate the role of \(w_{base}\) in mediating between prior knowledge and learned representations, demonstrating that an appropriate weight configuration effectively balances different performance objectives.

\begin{table}[htbp]
  \centering
  \caption{The Effect of Base Prediction Weight}
  \label{tab:w_base}
  \begin{tabular}{lcccc}
  \toprule\multirow{2}{*}{$w_{base}$} & \multicolumn{2}{c}{Before} & \multicolumn{2}{c}{After}\\
    \cmidrule(lr){2-3} \cmidrule(lr){4-5} 
    & Top-1 $\uparrow$ & Top-5$\uparrow$ & Top-1 $\uparrow$ & Top-5 $\uparrow$\\
    \midrule
    0.0   & 53.00\% & 79.80\% & 75.00\% & \textbf{89.40}\%\\
    0.1   & 55.00\% & 80.20\% & 76.20\% & \textbf{89.40}\%\\
    0.2  & 55.20\% & \textbf{80.40}\% & 76.20\% & 88.60\%\\
    0.3  & \textbf{55.40}\% & \textbf{80.40}\%  & \textbf{76.40}\% & 88.80\%\\
    0.4  & \textbf{55.40}\% & \textbf{80.40}\% & \textbf{76.40}\% & 88.80\%\\
    \bottomrule
  \end{tabular}
\end{table}


\subsubsection{The Effect of Hamming Weight Scaling Factor}
The Hamming weight scaling factor $w_{{hamming}} = 5$ calibrates the influence of Hamming distance measurements in our framework. As shown in Table~\ref{tab:w_hamming}, this parameter balances initial selection performance with optimized results.

The introduction of $w_{hamming}$ addresses a key challenge: properly scaling Hamming distance contribution to align with optimization objectives. At $w_{hamming} = 5.0$, we observe optimal Top-1 attack accuracy after optimization (76.20\%) while maintaining competitive Top-5 performance (88.60\%).

The parameter demonstrates a clear trade-off: lower values yield weaker initial performance, while higher values improve initial metrics but reduce optimization potential. The selected $w_{hamming} = 5.0$ represents the optimal balance where Hamming distance guidance enhances rather than constrains the optimization process.

\begin{table}[htbp]
  \centering
  \caption{The Effect of Hamming Weight Scaling Factor}
  \label{tab:w_hamming}
  \begin{tabular}{lccccc}
  \toprule\multirow{2}{*}{$w_{hamming}$} & \multicolumn{2}{c}{Before} & \multicolumn{2}{c}{After}\\
    \cmidrule(lr){2-3} \cmidrule(lr){4-5} 
    & Top-1 $\uparrow$ & Top-5$\uparrow$ & Top-1 $\uparrow$ & Top-5 $\uparrow$\\
    \midrule
    0.0   & 53.00\% & 79.40\% & 75.00\% & 88.60\%\\
    2.5   & 54.20\% & 79.40\% & 74.20\% & 89.00\%\\
    5.0  & 55.20\% & 80.40\% & \textbf{76.20}\% & 88.60\%\\
    7.5  & \textbf{55.80}\% & \textbf{80.60}\%  & \textbf{76.20}\% & \textbf{89.20}\%\\
    10.0  & \textbf{55.80}\% & 80.40\% & 75.40\% & 88.60\%\\
    \bottomrule
  \end{tabular}
\end{table}

\subsubsection{The Effect of Learning Rate}
\begin{table}[htbp]
  \centering
    \small 
  \setlength{\tabcolsep}{4pt} 
  \caption{The Effect of Learning Rate}
  \label{tab:lr}
  \begin{tabular}{lccccc}
    \toprule {$lr$}& Top-1 $\uparrow$ & Top-5  $\uparrow$ & KNN dist.  $\downarrow$ & mAP  $\uparrow$ & $S_{attack}$  $\uparrow$\\
    \midrule
    0.0005    & 63.20\% & 83.80\% & 1129.75 & 83.66\% & 0.5268\\
    0.0010    & 71.00\% & 87.60\% & 1089.18 & 87.45\% & 0.5713\\
    0.0015  & 76.20\% & 88.60\% & 1082.25 & \textbf{89.54}\% & 0.6066\\
    0.0020  & \textbf{77.80}\% & \textbf{90.20}\% & \textbf{1076.04} & 88.55\% & 0.6093\\
    0.0025  & 77.60\% & 89.40\% & 1080.43 & 88.16\% & \textbf{0.6144}\\
    \bottomrule
  \end{tabular}
\end{table}
\begin{figure}[htbp]
  \centering
  \includegraphics[width=0.9\linewidth]{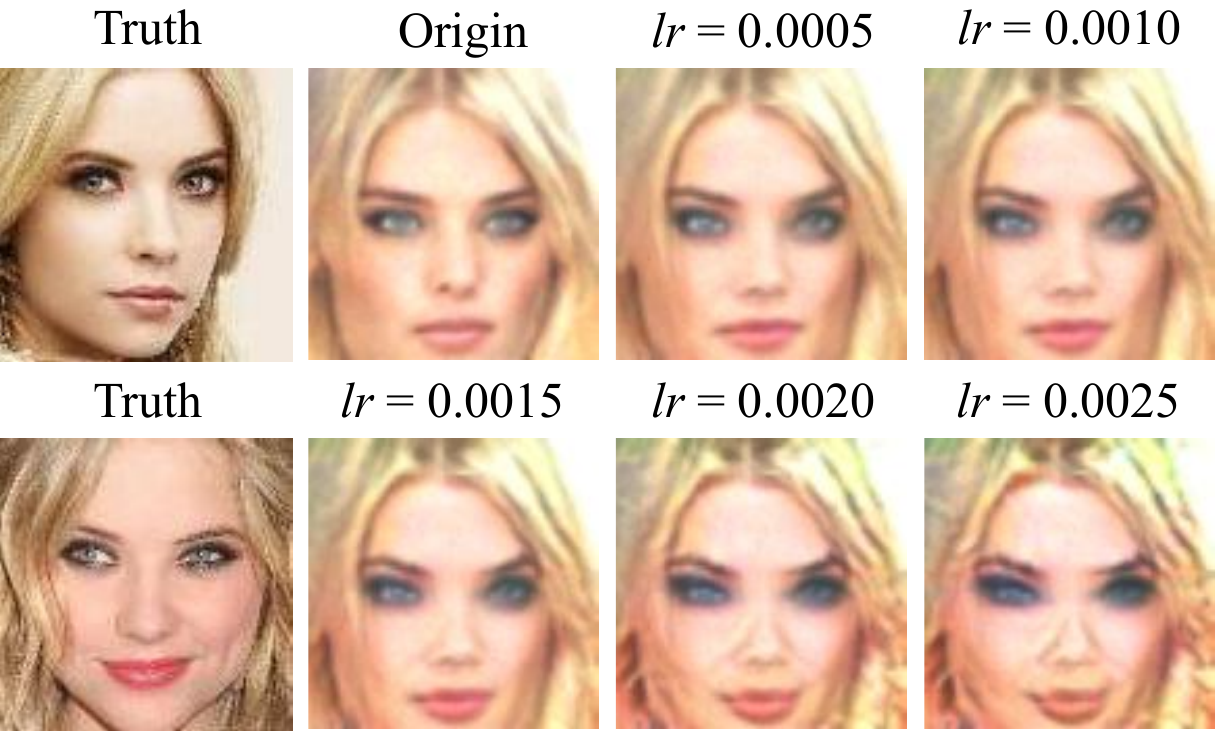}
  \caption{Learning Rate Effects on Image Reconstruction Quality}
   \label{fig:lr}
\end{figure}
Figure~\ref{fig:lr} provides a visual comparison of reconstructed images under different learning rates ($lr$), revealing a clear progression in optimization quality. Starting from the original not optimized (\textbf{Origin}), the reconstruction progressively incorporates more features of the training set (\textbf{Truth}) as $lr$ increases to 0.0015, demonstrating improved semantic fidelity and visual coherence. However, beyond this threshold at $lr$ values of 0.0020 and 0.0025, the images exhibit noticeable distortions, indicating over-optimization.

This visual trend is corroborated by the quantitative metrics in Table~\ref{tab:lr}. While higher learning rates ($lr$ = 0.0020-0.0025) achieve marginal improvements in certain metrics such as Top-1 attack accuracy and $S_{attack}$, they come at the cost of reduced mAP performance and visual quality degradation. The learning rate $lr$ = 0.0015 achieves the optimal balance, delivering the highest mAP (89.54\%) while maintaining competitive performance across other metrics and preserving visual integrity in the reconstructed images. This configuration was therefore selected as it optimally balances quantitative performance with qualitative reconstruction quality.


\subsubsection{The Effect of Intermediate Denoising Steps}

\begin{table}[htbp]
  \centering
    \small 
  \setlength{\tabcolsep}{4.5pt} 
  \caption{The Effect of Intermediate Denoising Steps}
  \label{tab:N}
  \begin{tabular}{lccccc}
    \toprule {$N$}& Top-1 $\uparrow$ & Top-5  $\uparrow$ & KNN dist.  $\downarrow$ & mAP  $\uparrow$ & $S_{attack}$  $\uparrow$\\
    \midrule
    300    & \textbf{76.60}\% & 88.60\% & \textbf{1075.06} & 87.44\% & 0.5976\\
    250    & 74.60\% & 88.60\% & 1079.98 & 88.49\% & 0.5987\\
    200  & 74.80\% & \textbf{89.00}\% & 1085.53 & 88.20\% & 0.6007\\
    150  & 75.40\% & 88.00\% & 1083.26 & 87.84\% & 0.6014\\
    100  & 76.20\% & 88.60\% & 1082.25 & \textbf{89.54}\% & \textbf{0.6066}\\
    50  & 72.60\% & 88.20\% & 1085.49 & 87.88\% & 0.5825\\
    10  & 63.80\% & 83.60\% & 1126.45 & 82.97\% & 0.5239\\
    \bottomrule
  \end{tabular}
\end{table}

Table~\ref{tab:N} illustrates how intermediate denoising steps $N$ affect model performance, revealing a complex relationship where different metrics peak at varying step counts and highlighting trade-offs in denoising granularity.

The attack adaptation metric $S_{attack}$ increases as $N$ decreases from 300 to 100, peaking at 0.6066 when $N=100$. Similarly, mAP reaches its maximum of 89.54\% at $N=100$, indicating optimal retrieval performance. However, Top-1 accuracy peaks at $N=300$ (76.60\%) with minimal KNN distance (1075.06), showing finer denoising benefits certain discriminative metrics.

The sharp performance drop at $N=10$ across all metrics demonstrates the necessity of sufficient denoising steps, with Top-1 accuracy decreasing by 12.80\% and $S_{attack}$ by 0.0827 compared to $N=100$. This confirms that inadequate denoising fails to properly reconstruct semantic features essential for effective hash center estimation.

The selected $N=100$ represents a practical compromise, delivering strong performance across multiple metrics while maintaining computational efficiency compared to higher step counts. This configuration prioritizes our key selection metric ($S_{attack}$) while maintaining competitive performance on other evaluation dimensions.

The parameter configurations adopted in our main experiments reflect a pragmatic approach aligned with real-world black-box scenarios. By selecting values that balance attack effectiveness ($S_{attack}$), attack accuracy, retrieval performance, computational efficiency, and visual quality—rather than optimizing for any single metric—we ensure robust performance under the constraints of practical deployment where exhaustive parameter tuning is infeasible.

\end{document}